\documentclass[final,1p,times,twocolumn]{elsarticle}
\usepackage[ruled,linesnumbered]{algorithm2e}
\SetKwRepeat{Do}{do}{while}%

\usepackage{amssymb}
\usepackage{amsmath}
\usepackage{float}

\journal{Robotics and Autonomous Systems}

\begin{document}

\begin{frontmatter}

%% Title, authors and addresses

%% use the tnoteref command within \title for footnotes;
%% use the tnotetext command for theassociated footnote;
%% use the fnref command within \author or \affiliation for footnotes;
%% use the fntext command for theassociated footnote;
%% use the corref command within \author for corresponding author footnotes;
%% use the cortext command for theassociated footnote;
%% use the ead command for the email address,
%% and the form \ead[url] for the home page:
%% \title{Title\tnoteref{label1}}
% \tnotetext[tnote 62]{This work was partially supported by the National Natural Science Foundation of China (62303486).}
%% \author{Name\corref{cor1}\fnref{label2}}
%% \ead{email address}
%% \ead[url]{home page}
%% \fntext[label2]{}
%% \cortext[cor1]{}
%% \affiliation{organization={},
%%             addressline={},
%%             city={},
%%             postcode={},
%%             state={},
%%             country={}}
%% \fntext[label3]{}

\title{Bridging Probabilistic Inference and Behavior Trees: An Interactive Framework for Adaptive Multi-Robot Cooperation}

%% use optional labels to link authors explicitly to addresses:
%% \author[label1,label2]{}
%% \affiliation[label1]{organization={},
%%             addressline={},
%%             city={},
%%             postcode={},
%%             state={},
%%             country={}}
%%
%% \affiliation[label2]{organization={},
%%             addressline={},
%%             city={},
%%             postcode={},
%%             state={},
%%             country={}}

\author[1]{Chaoran Wang \fnref{label2}}
\ead{chaoran_w@zju.edu.cn}

\affiliation[1]{organization={School of Aeronautic and Astronautics, Zhejiang University},%Department and Organization
            addressline={No. 866, Yuhangtang Road, Xihu District}, 
            city={Hangzhou},
            postcode={310027}, 
            state={Zhejiang},
            country={China}}
            
\affiliation[2]{organization={Shanghai Huawei Technologies Co., Ltd},%Department and Organization 
            city={Shanghai},
            postcode={201799}, 
            country={China}}

\author[2]{Jingyuan Sun}
\cortext[cor1]{Corresponding author}
\ead{sunjingyuan1@huawei.com}
\author[1]{Yanhui Zhang}

\author[1]{Changju Wu}

% Email id of the second author

%% Abstract
\begin{abstract}
This paper proposes an Interactive Inference Behavior Tree (IIBT) framework that integrates behavior trees (BTs) with active inference under the free energy principle for distributed multi-robot decision-making. The proposed IIBT node extends conventional BTs with probabilistic reasoning, enabling online joint planning and execution across multiple robots. It remains fully compatible with standard BT architectures, allowing seamless integration into existing multi-robot control systems. Within this framework, multi-robot cooperation is formulated as a free-energy minimization process, where each robot dynamically updates its preference matrix based on perceptual inputs and peer intentions, thereby achieving adaptive coordination in partially observable and dynamic environments. The proposed approach is validated through both simulation and real-world experiments, including a multi-robot maze navigation and a collaborative manipulation task, compared against traditional BTs(\textit{https://youtu.be/KX\_oT3IDTf4}). Experimental results demonstrate that the IIBT framework reduces BT node complexity by over 70\%, while maintaining robust, interpretable, and adaptive cooperative behavior under environmental uncertainty.\par
\end{abstract}

%%Graphical abstract
\begin{graphicalabstract}
\includegraphics[width=\textwidth]{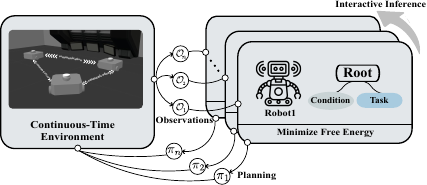}  
\end{graphicalabstract}

%%Research highlights
\begin{highlights}

\item A novel Interactive Inference Behavior Tree (IIBT) framework is proposed.

\item Integrates Active Inference with Behavior Trees for multi-robot decision-making.

\item Enables distributed and adaptive cooperation under uncertainty and partial observability.

\item Introduces joint preference matrices for inter-robot reasoning and policy alignment.

\item Validated through simulation and real-world multi-robot tasks, showing robustness gains.
\end{highlights}

%% Keywords
\begin{keyword}
%% keywords here, in the form: keyword \sep keyword

%% PACS codes here, in the form: \PACS code \sep code

%% MSC codes here, in the form: \MSC code \sep code
%% or \MSC[2008] code \sep code (2000 is the default)
Interactive inference, behavior tree, multirobot, joint action.
\end{keyword}

\end{frontmatter}

\section{Introduction}
\label{sec:introduction}
Cooperative decision-making among multiple robots operating in dynamic and uncertain environments remains a fundamental challenge in autonomous systems~\cite{lanillos2021active, priorelli2024embodied}. As task complexity grows—ranging from industrial assembly and warehouse logistics to search-and-rescue operations—robot teams must not only coordinate actions and share information but also adapt their behaviors to changing environmental conditions in real time~\cite{pezzulo2024active, friston2023federated}. Traditional centralized planning approaches provide global coordination capabilities but often suffer from high computational cost and limited scalability~\cite{maisto2023interactive, wirkuttis2021leading, friedman2021active, clodic2021implement}. Conversely, fully distributed or reactive control architectures offer rapid responses but frequently fail to maintain coherent team-level strategies under uncertainty.

To balance structured decision-making with real-time adaptability, BTs have emerged as a widely adopted control paradigm in both robotics and game AI. BTs offer a modular, hierarchical, and interpretable framework that enhances code reusability, debugging efficiency, and system transparency~\cite{ghallab2014actor's, nixon1999process}. Their execution semantics—based on ticking nodes that return \textit{Success}, \textit{Failure}, or \textit{Running} statuses—allow robots to reactively adapt to changing conditions without requiring a complete redesign of the control logic~\cite{venkata2023kt, hull2024communicating}. However, once a BT structure is constructed, it remains inherently deterministic~\cite{gugliermo2023learning}. This limitation makes it challenging to apply BTs in scenarios characterized by partial observability, dynamic task dependencies, and evolving cooperation requirements~\cite{colledanchise2019towards, luo2019importance}. As a result, BT-based systems often rely on static decision paths, which can degrade performance when environmental or task-related conditions deviate from design-time assumptions.

Meanwhile, interactive inference, grounded in the free energy principle, provides a probabilistic foundation for perception, prediction, and decision-making~\cite{kim2000multi, li2024multi}. By minimizing expected free energy, agents can iteratively infer hidden states and select action policies that balance exploratory information gathering with goal-directed behavior~\cite{pezzato2023active}. Despite its success in cognitive modeling and single-robot active perception, the application of interactive inference to multi-robot systems remains limited. More importantly, existing inference frameworks are typically monolithic or centralized, making them difficult to embed into modular, node-based decision architectures like BTs. Consequently, a fundamental gap persists between the interpretability and modularity of BTs and the probabilistic adaptability of inference-based methods.

\begin{figure}[t]
    \centering
    \includegraphics[width=0.8\textwidth]{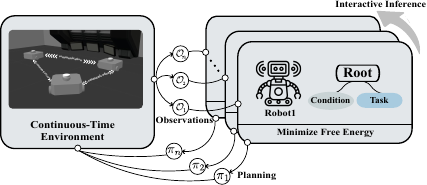}
    \caption{Overview of the proposed framework. Multiple robots perform interactive inference to jointly minimize free energy, dynamically update their policies, and coordinate actions in a shared, continuously evolving environment.}
    \label{fig1}
\end{figure}

To bridge this gap, this paper proposes an \textit{Interactive Inference Behavior Tree} (IIBT) framework that embeds free-energy–based probabilistic reasoning directly into BT execution nodes. The proposed framework preserves the modularity and interpretability of BTs while enabling each node to perform adaptive inference based on contextual observations. Through this integration, multiple robots can jointly infer and update their policies online, dynamically adapting to environmental variations and the actions of other agents during task execution.

The main contributions of this work are summarized as follows:
\begin{itemize}
    \item \textbf{Integration of probabilistic inference into BTs:} We propose a novel IIBT node that seamlessly integrates free-energy–based inference into BT execution semantics, enabling online adaptation without altering the BT structure.
    \item \textbf{Distributed cooperative policy selection:} Each node performs local inference based on expected free energy, supporting scalable, decentralized, and coherent decision-making among multiple robots.
    \item \textbf{Comprehensive experimental validation:} The proposed approach is evaluated through both simulation and real-world experiments, demonstrating significant improvements in task success rate, decision efficiency, and BT complexity compared to conventional approaches.
\end{itemize}

The remainder of this paper is organized as follows. Section~\ref{sec:related} reviews related work on interactive inference and BT-based planning. Section~\ref{sec:preliminary} formulates the cooperative decision-making problem and introduces the theoretical foundation. Section~\ref{sec:method} presents the proposed IIBT framework in detail. Implementation and case studies are described in Section~\ref{sec:implementation}, while Section~\ref{sec:experiments} reports experimental results. Finally, Section~\ref{sec:conclusion} discusses the findings and concludes the paper.
\section{Related Work}
\label{sec:related}

\subsection{Interactive Inference in Robotic Systems}

Interactive inference, grounded in the free energy principle, has emerged as a powerful paradigm for unified perception, prediction, and decision-making under uncertainty~\cite{friston2023federated, pezzulo2024active}. By formulating control as a process of minimizing expected free energy, agents can iteratively update their beliefs about hidden states and select policies that balance epistemic exploration with pragmatic goal-directed actions~\cite{lanillos2021active, pezzato2023active}. 

Initial research primarily focused on cognitive modeling and single-robot active perception~\cite{kim2000multi, li2024multi}. More recent studies have extended these ideas to multi-robot contexts, including distributed control~\cite{maisto2023interactive}, federated inference and belief sharing~\cite{friston2024federated}, and collective state estimation in partially observable environments~\cite{wakayama2024active}. Additionally, recent works have explored implicit coordination mechanisms where robots coordinate without explicit communication by inferring the latent intentions of teammates~\cite{bramblett2025implicit}. 

While these approaches demonstrate the versatility of active inference in robotics, most rely on centralized generative models or global state synchronization, which limit scalability in realistic multi-robot deployments. Furthermore, these methods typically lack structured representations for hierarchical task decomposition, which constrains their integration into modular decision-making architectures.

\subsection{Behavior Trees for Robotic Planning}

BTs have become a prominent alternative to classical decision architectures such as finite state machines, offering a modular, hierarchical, and interpretable control framework~\cite{colledanchise2021implementation}. By decomposing complex behaviors into control nodes (e.g., Sequence, Selector) and execution nodes (e.g., Condition, Action), BTs allow developers to build scalable decision policies that are reusable and easily debuggable~\cite{colledanchise2019towards, venkata2023kt}. 

BTs have been widely applied in robotic navigation~\cite{luo2019importance}, manipulation~\cite{gugliermo2023learning}, multi-robot coordination~\cite{clodic2021implement}, and human-robot collaboration~\cite{hull2024communicating}. To enhance adaptability, researchers have integrated BTs with machine learning~\cite{li2024embedding}, symbolic planning~\cite{liu2024autonomous}, and probabilistic models~\cite{scheide2025synthesizing}, as well as studied BT performance metrics and design evaluation methodologies~\cite{gugliermo2024evaluating}. 

Despite these advances, conventional BTs remain largely deterministic and static once defined. Most extensions focus on offline learning or external probabilistic reasoning layers, rather than incorporating probabilistic inference directly into BT execution semantics. As a result, current approaches still struggle to handle dynamic task priorities, partial observability, and emergent multi-robot interactions in a unified framework.

\subsection{Research Gap and Motivation}

In summary, two complementary research lines have emerged: inference-based methods offer probabilistic reasoning and adaptability but lack modular structure, while BT-based approaches provide interpretability and composability but cannot reason probabilistically or adapt online. A few attempts have combined active inference and BTs for reactive single-agent control~\cite{pezzato2023active}, but these efforts stop short of embedding free-energy–based reasoning \textit{within} BT execution nodes.

To the best of our knowledge, no prior work has integrated interactive inference directly into the execution semantics of BTs to enable distributed, adaptive, and cooperative multi-robot decision-making. This gap motivates the present work, which aims to unify these complementary strengths through the proposed \textit{Interactive Inference Behavior Tree (IIBT)} framework.

\section{Preliminary}
\label{sec:preliminary}

\subsection*{Notation}
To facilitate the following derivations, we summarize the main symbols and their definitions in Table~\ref{tab:notation}.

\begin{table}[h]
\centering
\caption{Notation summary used in the preliminary section.}
\label{tab:notation}
\begin{tabular}{ccl}
\hline
Symbol                                & \multicolumn{2}{c}{Description}                                                              \\ \hline
$\mathcal{R}_i$                       & \multicolumn{2}{c}{The $i$-th robot in a team of $N$ robots}                                 \\
$\mathcal{O}^i_{\tau}$ \&             & \multicolumn{2}{c}{Observation received by robot $i$ at time $\tau$}                         \\
$s^i$                            & \multicolumn{2}{c}{Hidden state of robot $\mathcal{R}_i$}                                 \\
$s^i_\tau$                            & \multicolumn{2}{c}{State of robot $i$ at time $\tau$}                                 \\
$\Pi^i = \{\pi^i_1, \dots, \pi^i_K\}$ & \multicolumn{2}{c}{Policy set available to robot $i$}                                        \\
$\pi^i_k$                             & \multicolumn{2}{c}{The $k$-th policy for robot $i$}                                          \\
$\pi_*^i$                             & \multicolumn{2}{c}{The optimal policy for robot $i$}                                         \\
$\pi_{\tau}^i$                        & \multicolumn{2}{c}{The policy for robot $i$ at time $\tau$}                                         \\
$\mathcal{P}$                         & \multicolumn{2}{c}{Generative distribution}                                                  \\
$\mathcal{Q}$                         & \multicolumn{2}{c}{Variational posterior distribution}                                       \\
$\mathcal{A}^i$                       & \multicolumn{2}{c}{Observation likelihood matrix $\mathcal{P}(\mathcal{O}^i_\tau|s^i_\tau)$} \\
$\mathcal{B}^i_{\pi}$                 & \multicolumn{2}{c}{State transition matrix $\mathcal{P}(s^i_\tau|s^i_{\tau-1}, \pi^i_k)$}    \\
$\mathcal{C}^i$                       & \multicolumn{2}{c}{Outcome preference prior}                                                 \\
$\mathcal{D}^i$                       & \multicolumn{2}{c}{Initial state prior $\mathcal{P}(s^i_1)$}                                 \\
$\mathcal{E}$                         & \multicolumn{2}{c}{Prior preference over policies}                                           \\
$\mathcal{F}(\pi^i_k)$                & \multicolumn{2}{c}{Variational free energy under policy $\pi^i_k$}                           \\
$\mathcal{G}(\pi^i_k)$                & \multicolumn{2}{c}{Expected free energy under policy $\pi^i_k$}                              \\
$\gamma$                              & \multicolumn{2}{c}{Precision parameter balancing planning and inference}                     \\
$\sigma(\cdot)$                       & \multicolumn{2}{c}{Softmax function}                                                         \\ \hline
\end{tabular}
\end{table}

\subsection{Generative Model for Multi-Robot Inference}

We consider a team of $N$ cooperative robots $\mathcal{R} = \{ \mathcal{R}_1, \dots, \mathcal{R}_N \}$ operating in a dynamic environment. Each robot $\mathcal{R}_i$ maintains an internal model of the environment through a generative process that relates hidden states, actions (policies), and observations over time. This model serves as the foundation for inference and planning.

The joint probability of the observation sequence $\mathcal{O}^i_{1:T}$, latent states $s^i_{1:T}$, and policy $\pi^i_k$ for robot $i$ can be expressed as:

\begin{equation}
\begin{split}
    \mathcal{P}\left (\mathcal{O}^{i}_{1:T},\bar{s}^{i}_{1:T}|\pi^{i} \right )&= \mathcal{P}\left ( s^{i}_{1}  \right )\prod_{\tau =1}^{T}\mathcal{P}\left(\mathcal{O}^{i} _{\tau}|s_{\tau}^{i}\right)\prod_{\tau =2}^{T} \mathcal{P}\left(s_{\tau}^{i}|s^{i}_{\tau-1},\pi^{i} \right) \\ 
    &=s^{i}_{1}\cdot \mathcal{D}^{i} \prod_{\tau=1}^{T}\mathcal{O}^{i}_{\tau}\cdot \mathcal{A}^{i}s_{\tau}^{i}\prod_{\tau=1}^{T}s_{\tau}^{i}\cdot \mathcal{B}^{i}s^{i}_{\tau-1} \label{eq:gen_model},
\end{split}
\end{equation}

Where the terms are interpreted as follows:
\begin{itemize}
    \item $\mathcal{P}(s^i_1)$: Prior over the initial hidden state, encoded by $\mathcal{D}^i$.
    \item $\mathcal{P}(\pi^i_k)$: Prior over policies, influenced by $\mathcal{E}$ and $\mathcal{C}^i$.
    \item $\mathcal{P}(\mathcal{O}^i_\tau | s^i_\tau)$: Likelihood of observation given the state, parameterized by $\mathcal{A}^i$.
    \item $\mathcal{P}(s^i_\tau | s^i_{\tau-1}, \pi^i_k)$: Transition model under policy $\pi^i_k$, parameterized by $\mathcal{B}^i_{\pi}$.
\end{itemize}

\begin{figure}[t]
    \centering
    \includegraphics[width=0.6\textwidth]{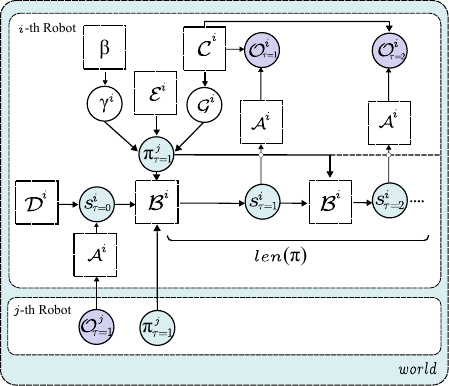}
    \caption{The figure illustrates the interactive inference process between robots $\mathcal{R}_{i}$ and $\mathcal{R}_{j}$ using a generative model.}
    \label{fig2}
\end{figure}

This generative structure forms the basis for both state estimation and action planning, capturing the causal relationships between actions, latent states, and observations in multi-robot collaboration.

\subsection{Variational Inference and Evidence Lower Bound}
\label{sec:elbo}

In probabilistic robotics and decision-making, computing the exact posterior distribution over hidden states and strategies, $\mathcal{P}(s^i_{1:T}, \pi^i_k | \mathcal{O}^i_{1:T})$, is intractable due to the exponential growth of the state space and the nonlinear observation models involved~\cite{blei2017variational,jordan1999introduction}. Variational inference provides a tractable approximation by introducing a surrogate distribution $\mathcal{Q}(s^i_{1:T}, \pi^i_k)$ and minimizing its Kullback-Leibler (KL) divergence from the true posterior:

\begin{equation}
\mathrm{KL}\left[ \mathcal{Q}(s^i_{1:T}, \pi^i_k) \parallel \mathcal{P}(s^i_{1:T}, \pi^i_k | \mathcal{O}^i_{1:T}) \right].
\end{equation}

This objective is equivalent to maximizing the \textit{Evidence Lower Bound} (ELBO), a standard formulation in Bayesian inference~\cite{kingma2014auto,blei2017variational}:

\begin{equation}
\mathcal{L} 
= 
\mathbb{E}_{\mathcal{Q}} \left[ \ln \mathcal{P}(\mathcal{O}^i_{1:T}, s^i_{1:T}, \pi^i_k) \right] 
- 
\mathbb{E}_{\mathcal{Q}} \left[ \ln \mathcal{Q}(s^i_{1:T}, \pi^i_k) \right].
\end{equation}

The negative ELBO is referred to as the \textit{variational free energy}~\cite{friston2010free,friston2015active}, which can be expressed as:

\begin{equation}
\mathcal{F}(\pi^i_k) 
= 
\mathrm{KL}\left[ \mathcal{Q}(s^i_{1:T}, \pi^i_k) \parallel \mathcal{P}(s^i_{1:T}, \pi^i_k | \mathcal{O}^i_{1:T}) \right] 
- 
\ln \mathcal{P}(\mathcal{O}^i_{1:T}).
\end{equation}

Since the marginal likelihood $\ln \mathcal{P}(\mathcal{O}^i_{1:T})$ is constant with respect to the optimization objective, minimizing $\mathcal{F}$ is equivalent to minimizing the KL divergence. Expanding the terms yields:

\begin{equation}
\mathcal{F}(\pi^i_k) = 
\underbrace{\mathrm{KL}\left[ \mathcal{Q}(s^i_\tau | \pi^i_k) \parallel \mathcal{P}(s^i_\tau | s^i_{\tau-1}, \pi^i_k) \right]}_{\text{Complexity}} 
- 
\underbrace{\mathbb{E}_{\mathcal{Q}} \left[ \ln \mathcal{P}(\mathcal{O}^i_\tau | s^i_\tau) \right]}_{\text{Accuracy}}.
\label{eq:free_energy}
\end{equation}

Here, the \textbf{complexity term} penalizes divergence between the posterior and the prior transition model, constraining the internal model's deviation from known dynamics. The \textbf{accuracy term}, in contrast, rewards beliefs that better explain the observed sensory data.

By minimizing the variational free energy, the robot continuously aligns its internal generative model with external observations, enabling robust state estimation and situational awareness~\cite{parr2019generalised,pezzulo2015active}. This mechanism provides the foundation for interactive decision-making and collaborative policy selection in multi-agent systems.

\subsection{Expected Free Energy and Policy Selection}
\label{sec:efe}

While the variational free energy $\mathcal{F}$ governs \textit{perception} by inferring latent states from current observations, decision-making in uncertain environments requires reasoning about \textit{future outcomes}. This is achieved by minimizing the \textit{expected free energy} (EFE) $\mathcal{G}$ for each candidate policy $\pi^i_k$~\cite{parr2019generalised}:

\begin{equation}
\mathcal{G}(\pi^i_k) = 
\underbrace{\mathrm{KL}\left[ \mathcal{Q}(\mathcal{O}^i_\tau | \pi^i_k) \parallel \mathcal{P}(\mathcal{O}^i_\tau) \right]}_{\text{Extrinsic value (goal alignment)}}-\underbrace{\mathbb{E}_{\mathcal{Q}} \left[ H(\mathcal{O}^i_\tau | s^i_\tau) \right]}_{\text{Intrinsic value (information gain)}}
\label{eq:expected_free_energy}
\end{equation}

The expected free energy can be interpreted as the sum of two complementary terms:

- \textbf{Extrinsic value:} The first term is the Kullback-Leibler divergence between predicted outcomes and prior preferences $\mathcal{C}^i$, which encourages policies that lead to outcomes consistent with task goals and desired states.
- \textbf{Intrinsic value:} The second term represents expected information gain about hidden states. Maximizing this term promotes epistemic exploration by selecting policies that reduce uncertainty, improving future state estimation.

Together, these two components balance \textit{goal-directed exploitation} and \textit{uncertainty-reducing exploration}~\cite{schwartenbeck2019computational, scheide2025synthesizing}, a property particularly important for multi-robot coordination where future contingencies cannot be exhaustively enumerated.

The posterior probability of executing a specific policy is then defined by a softmax distribution that integrates both variational and expected free energy terms:

\begin{equation}
p(\pi^i_k) = \sigma \left( \ln \mathcal{E}_k - \mathcal{F}(\pi^i_k) - \gamma \mathcal{G}(\pi^i_k) \right)
\label{eq:policy_prob}
\end{equation}

Where $\mathcal{E}_k$ denotes a prior over policies (often uniform), $\mathcal{F}$ encodes the current-state evidence, and $\mathcal{G}$ captures the expected future utility of a policy. The hyperparameter $\gamma$ regulates the relative precision of decision-making: a larger $\gamma$ emphasizes epistemic actions (information-seeking), while a smaller $\gamma$ biases decisions toward exploiting known rewards.

This formulation ensures that each robot selects strategies that jointly minimize epistemic uncertainty and maximize task-relevant outcomes, enabling coherent and adaptive multi-robot coordination even under partial observability and environmental uncertainty.

\section{Methodology}
\label{sec:method}

\subsection{Approach Overview and System Architecture}

To enable distributed, adaptive, and cooperative decision-making in multi-robot systems, we propose an \textit{Interactive Inference Behavior Tree} (IIBT) framework that tightly integrates probabilistic inference with the modular decision-making structure of BTs. Conventional BT-based methods typically rely on pre-defined control logic, which limits their adaptability and robustness under uncertainty. In contrast, the proposed IIBT architecture embeds inference capabilities directly within BT execution nodes, allowing robots to update their beliefs online, dynamically adjust decision priorities, and coordinate with teammates in partially observable environments.\par

\begin{figure}[htbp]
    \centering
    \includegraphics[width=0.6\textwidth]{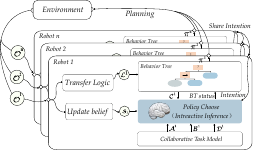}
    \caption{Workflow of interactive nodes in the BT. Each robot collects its local observations $\mathcal{O}^{i}_{\tau}$, abstracts them into logical variables $\mathcal{L}^{i}$, and updates its belief $s^{i}_{\tau}$. % <<< changed: added time subscript
    The BT emits a preference matrix $\mathcal{C}^{i}$ to the inference module, which queries task models for $\mathcal{A}^{i}, \mathcal{B}^{i}_{\pi}, \mathcal{D}^{i}$, % <<< changed: B^i -> B^i_{\pi}
    incorporates other robots’ intentions, and returns state/policy information back to the BT for execution.}
    \label{fig3}
\end{figure}

Each robot $\mathcal{R}_i$ maintains an internal generative model 
\[
\mathcal{M}^i = \{ \mathcal{A}^i, \mathcal{B}^i_{\pi}, \mathcal{C}^i, \mathcal{D}^i \},
\]
as defined in Table~\ref{tab:notation}, which captures the probabilistic relationships between sensory observations, latent states, action dynamics, and task preferences. Based on this model, the inference layer estimates latent state beliefs $s^i_{\tau}$, % <<< changed: add _{\tau}
predicts possible outcomes under candidate policies $\pi^i_k$, and evaluates their expected free energy. The resulting posterior beliefs are then fed into the BT layer, which orchestrates task execution through a structured hierarchy of condition and action nodes.

This integration transforms the BT from a static execution tree into a dynamic, belief-aware control architecture. It inherits the interpretability, modularity, and maintainability of BTs while gaining the adaptivity, robustness, and coordination capabilities associated with probabilistic inference. This hybrid design is particularly advantageous in multi-robot scenarios where agents must make decisions based on partial observations, dynamically changing objectives, and uncertain intentions communicated by peers.

\subsection{Interactive Inference BT Node (IIBT) Design}
The IIBT node serves as the core interface that bridges the reactive execution flow of a behavior tree with the probabilistic reasoning of the inference module. 
At each tick, the node not only decides which action policy to execute but also updates its belief and task preference in light of new observations and teammates’ inferred intentions. 
Algorithm~\ref{alg:strategy_selection} outlines the complete reasoning and execution cycle of the node for robot $\mathcal{R}_i$.

\begin{algorithm}[H]
\footnotesize
\caption{Strategy Selection Process within an IIBT Node for Robot $\mathcal{R}_i$}
\label{alg:strategy_selection}
\DontPrintSemicolon
\textit{Tick(}$\Im_{task}$\textit{)}\;
$\Im_{task}.r \gets \textit{running}$\;
Retrieve $\mathcal{M}^i=\{\mathcal{A}^i,\mathcal{B}^i_{\pi},\mathcal{C}^i,\mathcal{D}^i\}$ from task model\;
Retrieve preference matrix $\mathcal{C}^{i}$ from $\Im_{task}$\;
Acquire current observation $\mathcal{O}^{i}_{\tau}$ from environment\;
Update belief state $s^{i}_{\tau}$ and logic variables $\mathcal{L}^{i}$\; % <<< changed: add _{\tau}
Construct candidate policy set $\Pi^{i}$\;
\For{$k \in \{1, \dots, N\}\setminus i$}{
Receive $\mathcal{R}_{k}$ policy intention $\pi^{k}$\; % <<< changed: i -> k
$\Pi^{i} \gets \Pi^{i} \cup \{\pi^{k}\}$\; % <<< changed: overwrite -> union
}
$\pi^{i}_{\tau} \gets InteractiveInfer(s^{i}_{\tau}, \mathcal{M}^{i}, \Pi^{i})$\; % <<< changed: s^i -> s^i_{\tau}
\eIf{$\pi^{i}_{\tau} \equiv \pi^{i}_{stop}$}{
$\Im_{task}.r \gets success$\;
\textbf{return} $\Im_{task}.r$\;
}{
\While{$\mathcal{L}^{i} \notin \pi^{i}.prec$}{
$\mathcal{L}^{i}=f_{\mathcal{L}}(\pi^{i}.prec)$\;
$\mathcal{C}^{i} \gets \mathcal{C}^{i}+\mathcal{L}^{i}$\;
Reconstruct $\Pi^{i}_{task}$\;
\For{$k \in \{1, \dots, N\}\setminus i$}{
Receive $\mathcal{R}_{k}$ policy intention $\pi^{k}$\; % <<< changed: i -> k
$\Pi^{i} \gets \Pi^{i} \cup \{\pi^{k}\}$\; % <<< changed
}
$\pi^{i}_{\tau} \gets InteractiveInfer(s^{i}_{\tau}, \mathcal{M}^{i}, \Pi^{i},\mathcal{O}^{i}_{\tau})$\; % <<< changed
\If{Timeout}{
$\Im_{task}.r \gets failure$\;
\textbf{return} $\Im_{task}.r$\;
}
}
\eIf{$countConnectedRobot = N-1$}{
\texttt{Execute}($\pi^{i}_{\tau}$)\;
$\mathcal{C}^{i} \gets \mathcal{C}^{i}-\mathcal{L}^{i}$\;
}{
\texttt{Execute}($\pi^{i}_{wait}$)\;
}
\If{$\pi^{i}_{\tau} \in \pi^{i}_{stop}.prec$}{
$\pi^{i}_{\tau} \gets \pi^{i}_{stop}$\;
}
$\Im_{task}.r \gets running$\;
\textbf{return} $\Im_{task}.r$\;
}
\end{algorithm}

\paragraph{Role in the architecture.}
As shown in Fig.~\ref{fig3}, the IIBT node is the execution-time interface that closes the loop between probabilistic inference and the BT tick-cycle. At every tick, it (i) reads the current belief and observation, (ii) updates the preference matrix using symbolic logical evidence, (iii) incorporates peer intentions, (iv) runs interactive inference over the candidate policies, and (v) executes or defers actions depending on team connectivity and preconditions. The full tick routine is given in Alg.~\ref{alg:strategy_selection}.

As shown in Fig.~\ref{fig3}, the IIBT node closes the perception–inference–action loop during each BT tick. 
At runtime, it (i) collects local observations and current beliefs, (ii) transforms symbolic logic into quantitative preference updates, (iii) exchanges policy intentions with peers, and (iv) performs interactive inference to select the most plausible cooperative policy (Alg.~\ref{alg:strategy_selection}). 
This makes each node an autonomous reasoning unit that aligns its local execution policy with both environmental feedback and team-level belief consistency.

\paragraph{Inputs and maintained state.}
For robot $\mathcal{R}_i$, the node retrieves its generative model $\mathcal{M}^i=\{\mathcal{A}^i,\mathcal{B}^i_{\pi},\mathcal{C}^i,\mathcal{D}^i\}$ (line~3), obtains sensory observation $\mathcal{O}^{i}_{\tau}$ (line~5), and updates both the filtered belief $s^i_{\tau}$ and its symbolic abstraction $\mathcal{L}^i$ (line~6). % <<< changed: add _{\tau}
The candidate policy pool $\Pi^i$ (line~7) is continuously expanded with peer intentions $\pi^k$ from other robots (lines~8–11), creating a belief-aware and context-sensitive action space.

\paragraph{Model Semantics.}

Each robot maintains a latent state vector
\[
s^i_\tau = [s^{i}_{\text{loc}}, s^{i}_{\text{hold}}, s^{i}_{\text{place}}, s^{i}_{\text{free}}]^\top, % <<< changed: add ^i to components
\]
representing the robot’s belief in reaching a location, grasping, placing, or being idle. 
The inference process operates on the joint state
\[
s^{\text{joint}}_\tau = [\,s^{1}_\tau, s^{2}_\tau, \dots, s^{N}_\tau, s^{\text{result}}_\tau]^\top, % <<< changed: was s^i = [...]
\]
corresponding respectively to the robots’ beliefs and the global task outcome.  
Each state dimension is normalized as a probability distribution representing belief strength (e.g., $s^{i}_{\text{hold}} = [0.9,0.1]^\top$ indicates 90\% confidence of holding the object).

Here, $s^{\text{result}}_\tau$ is a global latent variable summarizing the cooperative task outcome (e.g., overall success, failure, or pending status). 
This concatenated representation allows the inference process to capture cross-robot dependencies such as temporal ordering, spatial coupling, and shared resource constraints. 
In practice, each IIBT node maintains and updates its own marginal belief $s^{i}_\tau$ at time $\tau$, 
but exchanges summarized information about other agents’ inferred states during coordination, 
thereby realizing a distributed yet coherent joint inference process across the team.

This formulation captures inter-robot dependencies within a unified generative process Eq.(\ref{eq:gen_model}).

\paragraph{Observation representation and joint observation matrix.}

For a system consisting of $N$ cooperative robots, the overall observation at discrete time step $\tau$ is represented as
\[
\mathcal{O}^{\text{joint}}_{\tau} = [\mathcal{O}^{1}_{\tau}, \mathcal{O}^{2}_{\tau}, \dots, \mathcal{O}^{N}_{\tau}, \mathcal{O}^{\text{result}}_{\tau}]^{\top}, % <<< changed: O^i_tau -> O^{joint}_tau
\]
where $\mathcal{O}^{\text{joint}}_{\tau}$ denotes the \textit{joint observation vector} of the multi-robot system.  
It concatenates all individual robots’ local observations and a task-level outcome vector, forming the sensory interface between the physical environment and the inference process.  
Each element of $\mathcal{O}^{i}_{\tau}$ is a binary or probabilistic indicator that reflects whether a certain physical or symbolic event is currently observed. % <<< clarified: O^i_tau is local

For each robot $\mathcal{R}_i$ $(i=1,\dots,N)$, the local observation vector
\[
\mathcal{O}^{i}_{\tau} = [\,o^{i}_1, o^{i}_2, \dots, o^{i}_{m_i}\,]^{\top} % <<< changed: ri -> i, m -> m_i
\]
encodes the status of its task-relevant latent variables.  
Each component $o^{i}_j$ $(j=1,\dots,m_i)$ is a binary observation associated with the $j$-th latent state $s^i_j$:
\[
o^{i}_j =
\begin{cases}
1, & \text{if the state } s^i_j \text{ is achieved at time } \tau,\\
0, & \text{otherwise.}
\end{cases}
\]
Hence, $\mathcal{O}^i_{\tau}$ provides a direct logical mapping from the robot’s perception space to its latent belief state $s^i_{\tau}$.  
In practice, these elements can be computed from sensor feedback, symbolic condition checks, or communication messages.  
For instance, if robot $\mathcal{R}_1$ successfully grasps an object, then $o^{1}_{\text{hold}}=1$ while other entries remain zero.

In addition to local observations, the system maintains a task-level observation vector $\mathcal{O}^{\text{result}}_{\tau}$ that summarizes the global cooperative outcome at the current time step:
\[
\mathcal{O}^{\text{result}}_{\tau} = [\,o^{\text{result}}_{\text{success}},\, o^{\text{result}}_{\text{failure}},\, o^{\text{result}}_{\text{null}}\,]^{\top}.
\]
Each entry in $\mathcal{O}^{\text{result}}_{\tau}$ is a binary indicator specifying whether the global task has reached a corresponding terminal or intermediate state:
\[
o^{\text{result}}_{k} =
\begin{cases}
1, & \text{if the overall task is in status } k \text{ at time } \tau,\\
0, & \text{otherwise.}
\end{cases}
\]
This vector serves as a shared global signal that allows all robots to condition their inference on collective task progress, enabling synchronization and cooperative adaptation among team members.

By vertically stacking all local and global observations, the joint observation matrix can be expressed as
\[
\mathcal{O}^{\text{joint}}_{\tau} = 
\begin{bmatrix}
o^{1}_{\text{loc}} & o^{1}_{\text{hold}} & o^{1}_{\text{place}} & o^{1}_{\text{free}} \\
o^{2}_{\text{loc}} & o^{2}_{\text{hold}} & o^{2}_{\text{place}} & o^{2}_{\text{free}} \\
\vdots & \vdots & \vdots & \vdots \\
o^{\text{result}}_{\text{success}} & o^{\text{result}}_{\text{failure}} & o^{\text{result}}_{\text{null}} & \text{--}
\end{bmatrix}^{\top},
\]
where “--” indicates a null or non-applicable entry.  
Each row corresponds to one agent (including the global result layer), while each column denotes a semantic dimension of the task space, such as localization, grasping, placement, or idle status.  
This joint structure allows the inference module to integrate heterogeneous sensory and symbolic information across agents, providing a unified observation basis for distributed belief updating within the IIBT framework.

\paragraph{Observation and transition model matrices.}

Consequently, the likelihood matrix $\mathcal{A}^{i}$ and transition matrix $\mathcal{B}^{i}_{\pi}$ are constructed in block-diagonal form to represent both individual robot dynamics and inter-robot dependencies:
\[
\mathcal{A}^{i} =
\begin{bmatrix}
\mathcal{A}^{i}_{r1} & 0 & 0 & \cdots & 0 \\
0 & \mathcal{A}^{i}_{r2} & 0 & \cdots & 0 \\
\vdots & \vdots & \vdots & \ddots & \vdots \\
0 & 0 & 0 & \mathcal{A}^{i}_{rN} & 0\\
0 & 0 & 0 & \cdots & \mathcal{A}^{i}_{\text{result}}
\end{bmatrix}, \qquad
\]\par
\[
\mathcal{B}^{i}_{\pi} =
\begin{bmatrix}
\mathcal{B}^{11} & \mathcal{B}^{12} & \cdots & \mathcal{B}^{1N} \\
\mathcal{B}^{21} & \mathcal{B}^{22} & \cdots & \mathcal{B}^{2N} \\
\vdots & \vdots & \ddots & \vdots \\
\mathcal{B}^{N1} & \mathcal{B}^{N2} & \cdots & \mathcal{B}^{NN}
\end{bmatrix}.
\]
Here, each block $\mathcal{A}^{i}_{rj}$ and $\mathcal{B}^{ij}$ corresponds to robot $\mathcal{R}_j$’s local observation and transition model, respectively. % <<< changed: clarify rj/j
The diagonal terms $\mathcal{B}^{ii}$ encode the self-dynamics of each robot, while the off-diagonal terms $\mathcal{B}^{ij}$ ($i \neq j$) represent the coupling effects between robots, such as physical interference, task dependencies, or coordination constraints.  
The bottom-right term $\mathcal{A}^{i}_{\text{result}}$ describes the observation likelihood of the global task outcome, associated with the global observation vector $\mathcal{O}^{\text{result}}_{\tau}$ defined earlier.

This block-structured formulation enables the inference module to reason about multi-agent dependencies probabilistically, while each IIBT node still performs local updates through decentralized message passing.

\paragraph{Local likelihood model.}

For each robot $\mathcal{R}_i$, the observation (likelihood) matrix $\mathcal{A}^i$ defines the conditional probability of receiving a particular observation given the current latent state:
\[
\mathcal{P}(\mathcal{O}^i_{\tau} \,|\, s^i_{\tau}) = \mathcal{A}^i s^i_{\tau},
\]
where $\mathcal{A}^i \in \mathbb{R}^{m_i \times n_i}$ maps the $n_i$-dimensional latent state space to the $m_i$-dimensional observation space.  
Each column of $\mathcal{A}^i$ specifies the likelihood distribution over observable outcomes when the system is in a particular hidden state.  

Physically, $\mathcal{A}^i$ captures the reliability of the perception or sensing process.  
For instance, a diagonal entry of $0.9$ indicates that the vision or gripper sensor correctly reflects the true state 90\% of the time, while the remaining $0.1$ models observation noise caused by occlusion, lighting change, or sensor failure.  
When the task involves symbolic communication (e.g., “object placed”), the same formulation applies, treating message acknowledgment as an observation channel.

\paragraph{Local transition model.}

The state-transition matrix $\mathcal{B}^i_{\pi}$ encodes how each robot’s internal belief evolves over time given its executed policy $\pi^i$.  
Formally,
\[
\mathcal{P}(s^i_{\tau+1} \,|\, s^i_\tau, \pi^i) = \mathcal{B}^i_{\pi^i} s^i_\tau,
\]
where $\mathcal{B}^i_{\pi^i} \in \mathbb{R}^{n_i \times n_i}$ is the action-specific transition matrix associated with policy $\pi^i$.  
Each column of $\mathcal{B}^i_{\pi^i}$ represents the probability distribution of the next state given the current state and action.  

Different actions correspond to different transition matrices:
\[
\mathcal{B}_{\text{move}},\; \mathcal{B}_{\text{pick}},\; \mathcal{B}_{\text{place}},\; \mathcal{B}_{\text{idle}},
\]
where, for example, $\mathcal{B}_{\text{move}}$ increases the probability of $s^{i}_{\text{loc}}=1$ as navigation proceeds,  
$\mathcal{B}_{\text{pick}}$ increases the probability of $s^{i}_{\text{hold}}=1$ after a successful grasp,  
$\mathcal{B}_{\text{place}}$ increases the probability of $s^{i}_{\text{place}}=1$ once placement is achieved,  
and $\mathcal{B}_{\text{idle}} \approx I_4$ maintains the current belief state when no action is executed.  

These transition probabilities are empirically estimated from execution logs, using success/failure ratios or temporal statistics, and are normalized column-wise to ensure valid probability distributions.  
By combining $\mathcal{A}^i$ and $\mathcal{B}^i_{\pi^i}$, each robot maintains a physically grounded generative model that connects sensory uncertainty, action dynamics, and latent belief updates within the Interactive Inference Behavior Tree framework.

\paragraph{Prior state distribution.}

The prior matrix $\mathcal{D}^i$ defines the initial belief over the latent task states of robot $\mathcal{R}_i$ before task execution begins.  
Formally, $\mathcal{D}^i \in \mathbb{R}^{n_i}$ is a column vector representing the initial probability distribution of the latent state $s^i_{\tau=0}$:
\[
\mathcal{P}(s^i_{\tau=0}) = \mathcal{D}^i,
\]
where $n_i$ denotes the number of task-relevant hidden states maintained by robot $\mathcal{R}_i$.  
Each element $d^i_j$ of $\mathcal{D}^i = [d^i_1, d^i_2, \dots, d^i_{n_i}]^\top$ specifies the prior probability of the $j$-th latent state being true at the initial time step. 
This initialization encodes the assumption that the robot has not yet reached a target location, grasped an object, or completed a placement at time $\tau = 0$.  

From a probabilistic perspective, $\mathcal{D}^i$ serves as the starting point of the generative process, defining the prior belief $\mathcal{P}(s^i_0)$ in the joint distribution Eq.(\ref{eq:gen_model}), where $\mathcal{P}(s^i_0)$ corresponds exactly to $\mathcal{D}^i$.  
This ensures that all subsequent inferences and belief updates within the IIBT framework remain grounded in a consistent probabilistic prior, promoting stable initialization and reproducible behavior across different robot agents.

The preference matrix $\mathcal{C}^{i}$ defines the extrinsic desirability of potential observation outcomes, guiding each robot’s policy selection toward goal-consistent behaviors.  
In the multi-robot setup, $\mathcal{C}^{i}$ is modeled jointly rather than independently per agent to ensure that all preferences are aligned with both individual objectives and collective task constraints.  
Formally, the joint preference matrix at time step $\tau$ is expressed as
\[
\mathcal{C}^{i} = [\,\mathcal{C}^{i}_{r1},\; \mathcal{C}^{i}_{r2},\; \dots,\; \mathcal{C}^{i}_{rN},\; \mathcal{C}^{i}_{\text{result}}\,]^{\top},
\]
where each block $\mathcal{C}^{i}_{rj} \in \mathbb{R}^{m_j}$ encodes the preference distribution of robot $\mathcal{R}_j$ over its local observation space $\mathcal{O}^j_{\tau}$, and $\mathcal{C}^{i}_{\text{result}}$ represents the team-level preference over global outcomes (e.g., task success or failure).  % <<< changed: "column" -> "block", O^i_tau -> O^j_tau

Each preference vector is defined as
\[
\mathcal{C}^{i}_{rj} = [\,c^{i}_{1},\, c^{i}_{2},\, \dots,\, c^{i}_{m_j}\,]^{\top},
\]
where $c^{i}_{k}$ denotes the desirability of observing outcome $o^j_k$ under the current task objective.  
High preference values bias the inference process toward policies expected to produce those outcomes, as reflected in the expected free energy computation of the Interactive Inference step.

The preference matrix $\mathcal{C}^i$ specifies desirable observation outcomes derived from task goals, e.g.,
high weight on $l_{\text{hold}}$ during grasping phases and on $l_{\text{place}}$ during assembly. During execution, $\mathcal{C}^i$ is adaptively updated through the logic-to-preference mapping $f_{\mathcal{L}}$, which injects symbolic conditions (e.g., preconditions or unmet goals) into the preference vector to bias the expected free energy.

Overall, the combination $\{\mathcal{A}^i,\mathcal{B}_{\pi}^i,\mathcal{C}^i,\mathcal{D}^i\}$ defines a compact yet physically grounded generative model that links sensory uncertainty, action dynamics, and goal preference. These matrices can be tuned directly from empirical robot data or analytically set based on system reliability, providing a clear interface between probabilistic inference and the robot’s physical control domain.

\paragraph{Tick-cycle flow.}
At the beginning of each tick the node sets its return status to \textit{running} (line~2) and constructs a local candidate policy pool $\Pi^i$ (line~7). It then collects the most recent policy intentions broadcast by other robots (lines~8–11) and augments $\Pi^i$ accordingly, yielding a peer-aware candidate set. Given $(s^{i}_{\tau},\mathcal{M}^{i},\Pi^{i})$, the node calls \textsc{InteractiveInfer} (line~12) to obtain the current policy $\pi^i_{\tau}$ via the free-energy based posterior (cf. Sec.~\ref{sec:efe}). If the selected policy is the terminal policy $\pi^i_{stop}$ (line~13), the node returns \textit{success} (lines~14–15). % <<< changed: s^i -> s^i_{\tau}

\begin{algorithm}[H]
\footnotesize
\caption{\textsc{InteractiveInfer (lite)}: One-step Active-Inference Scoring for $\mathcal{R}_i$}
\label{alg:interactive_infer_lite}
\DontPrintSemicolon
\KwIn{Belief $s^{i}_{\tau}$, model $\mathcal{M}^i=\{\mathcal{A}^i,\mathcal{B}^i_{\pi},\mathcal{C}^i,\mathcal{D}^i\}$, candidates $\Pi^{i}$, current observation $\mathcal{O}^{i}_{\tau}$, policy prior $\mathcal{E}$, precision $\gamma$} % <<< changed: s^i -> s^i_{\tau}
\KwOut{Selected policy $\pi^{i}_{\tau}$ and posterior $\mathcal{Q}(\pi^{i})$}
\BlankLine
$\mathcal{P}(\mathcal{O}^{i}_{\tau}) \gets \textsc{Softmax}(\mathcal{C}^{i})$ \tcp*{preference-induced outcome prior}
\ForEach{$\pi \in \Pi^{i}$}{
    $b' \gets \mathcal{B}^{i}(\cdot\,|\,\pi)\; s^{i}_{\tau}$ \tcp*{next-belief (one-step rollout)} % <<< changed
    $q(\mathcal{O}^{i}_{\tau}) \gets \mathcal{A}^{i}\; b'$ \tcp*{predicted outcome marginal}
    $E_{\mathrm{ext}} \gets D_{\mathrm{KL}}\!\big(q(\mathcal{O}^{i}_{\tau})\,\|\,\mathcal{P}(\mathcal{O}^{i}_{\tau})\big)$ \;
    $E_{\mathrm{int}} \gets \sum_{s} b'(s)\, H\!\big(\mathcal{A}^{i}(:,s)\big)$ \;
    $\mathcal{G}(\pi) \gets E_{\mathrm{ext}} - E_{\mathrm{int}}$ \;
    $\mathcal{F}(\pi) \gets - \sum_{s} s^{i}_{\tau}(s)\,\log \mathcal{A}^{i}\!\big(\mathcal{O}^{i}_{\tau}, s\big)$ \; % <<< changed
    $\textsc{Score}[\pi] \gets \ln \mathcal{E}(\pi) - \mathcal{F}(\pi) - \gamma\,\mathcal{G}(\pi)$ \;
}
$\mathcal{Q}(\pi^{i}) \gets \textsc{Softmax}\big(\textsc{Score}[\cdot]\big)$ \tcp*{log-sum-exp stabilized}
$\pi^{i}_{\tau} \gets \arg\max_{\pi \in \Pi^{i}} \mathcal{Q}(\pi^{i}=\pi)$ \;
\Return{$\pi^{i}_{\tau}$, $\mathcal{Q}(\pi^{i})$}
\end{algorithm}

\paragraph{Logical-to-preference shaping.}

In our multi-robot setup, the preference matrix $\mathcal{C}^{i}$ is also modeled jointly rather than independently per agent. 
Specifically, $\mathcal{C}^{i}$ is structured as a column-wise concatenation $\mathcal{C}^{i}= [\mathcal{C}^{i}_{r1}, \mathcal{C}^{i}_{r2},\dots, \mathcal{C}^{i}_{\text{result}}]^{\top}$,
where each block encodes the extrinsic preferences of one robot. 
$\mathcal{C}^{1}$ and $\mathcal{C}^{2}$ represent, respectively, the desired observation likelihoods for robots $\mathcal{R}_1$ and $\mathcal{R}_2$. 
Each element $c^{i}_{k}$ in $\mathcal{C}^{i}_{rj}=[c^{i}_{0},c^{i}_{1},\dots,c^{i}_{M}]^{\top}$ corresponds to the desirability of observing outcome $o$ under robot $\mathcal{R}_i$’s current task objective.

If preconditions for $\pi^{i}_{\tau}$ are not satisfied ($\mathcal{L}^{i}\notin\pi^{i}_{\tau}.\text{prec}$, line~17), the node computes the minimal logical evidence required to satisfy the precondition using the mapping $f_{\mathcal{L}}(\cdot)$ (line~18), and applies it as an additive update on $\mathcal{C}^{i}$ (line~19):
\[
\mathcal{L}^{i}=f_{\mathcal{L}}(\pi^{i}_{\tau}.\text{prec}),\quad
\mathcal{C}^{i}\leftarrow\mathcal{C}^{i}+\mathcal{L}^{i}.
\]
Intuitively, this raises the extrinsic preference for outcomes that make the precondition true, thereby biasing the EFE toward prerequisite-achieving actions. The node then reconstructs the task-specific policy pool $\Pi^{i}_{task}$ (line~20), refreshes peer intentions (lines~21–24), and re-runs \textsc{InteractiveInfer} (line~25). If a timeout occurs (lines~26–29), the node fails fast, returning \textit{failure}.

\paragraph{Execution and synchronization.}
When the communication layer reports that all teammates are connected ($countConnectedRobot=N-1$, line~31), the node executes the selected policy $\pi^{i}_{\tau}$ (line~32). Immediately after dispatch, it \emph{rolls back} the temporary preference boost associated with the just-satisfied logical increment (line~33), i.e.,
\[
\mathcal{C}^{i}\leftarrow \mathcal{C}^{i}-\mathcal{L}^{i},
\]
restoring the baseline preferences to avoid long-term drift. If full connectivity is not met, the node executes a wait policy $\pi^{i}_{wait}$ (line~35), preserving safety and coordination while messages converge.

\paragraph{Termination guard.}
If the currently selected policy becomes a member of the stop precondition set (line~37), the node promotes it to $\pi^{i}_{stop}$ (line~38). Otherwise, it continues in \textit{running} state (line~40) and returns control to the parent BT composite (line~41).

\paragraph{Discussion and interface}
The IIBT node exposes two light-weight interfaces to the rest of the BT: (i) a \emph{logic-to-preference} adapter $f_{\mathcal{L}}$ that transforms symbolic BT conditions into additive updates on $\mathcal{C}^i$, and (ii) the \textsc{InteractiveInfer} call that converts $(s^{i}_{\tau},\mathcal{M}^{i},\Pi^{i})$ into a softmax posterior over policies, using the variational free energy $\mathcal{F}$ (perception term) and expected free energy $\mathcal{G}$ (prospection term) defined in Secs.~\ref{sec:efe}. % <<< changed: s^i -> s^i_{\tau}
This design preserves BT interpretability and reactivity while endowing each node with uncertainty-aware, preference-driven adaptation.

\section{Implementation of Interactive Inference Nodes}
\label{sec:implementation}

\subsection{Robots Interactive Inference: A Simple Example}

This section presents a case study on interactive inference in a multirobot system, showing how robots plan behaviors under unknown objectives while minimizing free energy.\par

\begin{figure}[h]
    \centering
    \includegraphics[width=0.4\textwidth]{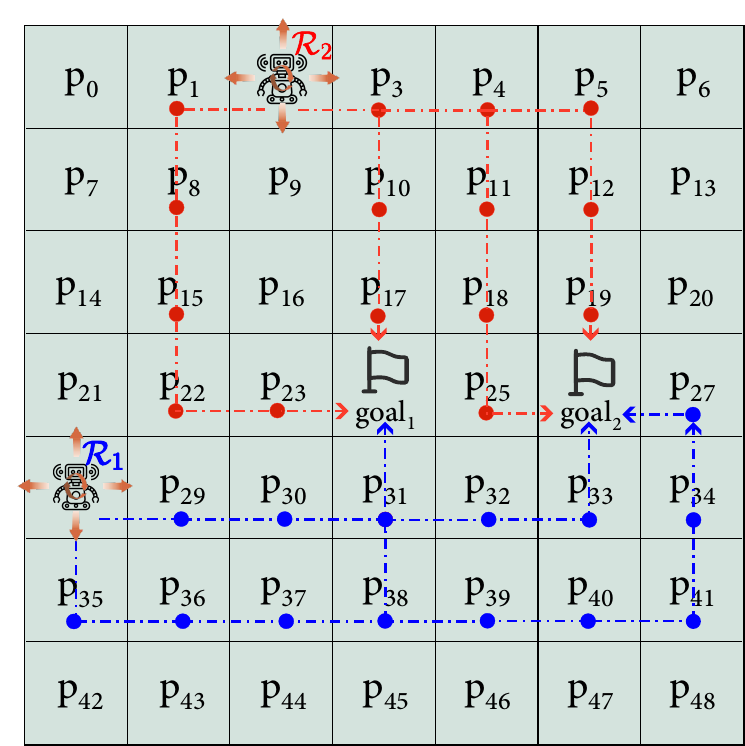}
    \caption{Two robots, $\mathcal{R}_{1}$ and $\mathcal{R}_{2}$, operate in a 7x7 grid, with cell positions labeled as $\{ p_{0}, p_{1}, \dots, p_{48} \}$. The environment features two goals, $goal_{1}$ and $goal_{2}$, and each robot creates paths to both goals.}
    \label{fig6}
\end{figure}

We examine two robots, $\mathcal{R}_{1}$ and $\mathcal{R}_{2}$, in an environment shown in Fig.~\ref{fig6} with two goals (\(goal_{1}\) and \(goal_{2}\)). The robots cannot identify their goals but can see each other's positions on a grid map with locations $\{p_{0}, p_{1}, \dots, p_{48}\}$.\par % <<< changed: p_2 -> p_1, notation fix

The strategy set \(\Pi^{1}\) includes the strategies for robot \(\mathcal{R}_{1}\) to achieve two goals. The paths to \(goal_{1}\) are \(\{ p_{28}, p_{29}, p_{30}, p_{31}, p_{24} \}\) and \(\{ p_{28}, p_{35}, p_{36}, p_{37}, p_{38}, p_{31}, p_{24} \}\). The paths to \(goal_{2}\) are \(\{ p_{28}, p_{29}, p_{30}, p_{31}, p_{32}, p_{33}, p_{26} \}\) and \(\{ p_{28}, p_{35}, p_{36}, p_{37}, p_{38}, p_{39}, p_{40}, p_{41}, p_{34}, p_{27}, p_{26} \}\). The strategy set for robot \(\mathcal{R}_{2}\) (\(\Pi^{2}\)) includes paths to \(goal_{1}\): \(\{ p_{2}, p_{1}, p_{8}, p_{15}, p_{22}, p_{23}, p_{24} \}\) and \(\{ p_{2}, p_{3}, p_{10}, p_{17}, p_{24} \}\); and paths to \(goal_{2}\): \(\{ p_{2}, p_{3}, p_{4}, p_{11}, p_{18}, p_{25}, p_{26} \}\) and \(\{ p_{2}, p_{3}, p_{4}, p_{5}, p_{12}, p_{19}, p_{26} \}\). % <<< fixed punctuation and colon consistency

The task requires robots \(\mathcal{R}_{1}\) and \(\mathcal{R}_{2}\) to reach different goals simultaneously. The combination of strategy selections is defined as \( \Pi^{1} \times \Pi^{2} = \{ (\pi^{1}, \pi^{2}) \mid \pi^{1} \in \Pi^{1}, \pi^{2} \in \Pi^{2} \} \). % <<< changed: added superscripts for consistency

For robot \(\mathcal{R}_{1}\), the hidden states are \(s^{1} = \{ s^{1}_{\tau}, s^{2}_{\tau} \}\), where \(s^{1}_{\tau} = [p_{0}, p_{1}, \ldots, p_{48}]^{\top}\) and \(s^{2}_{\tau} = [p_{0}, p_{1}, \ldots, p_{48}]^{\top}\). % <<< added ^\top and made consistent with previous section

\begin{figure}[h]
    \centering
    \includegraphics[width=0.7\textwidth]{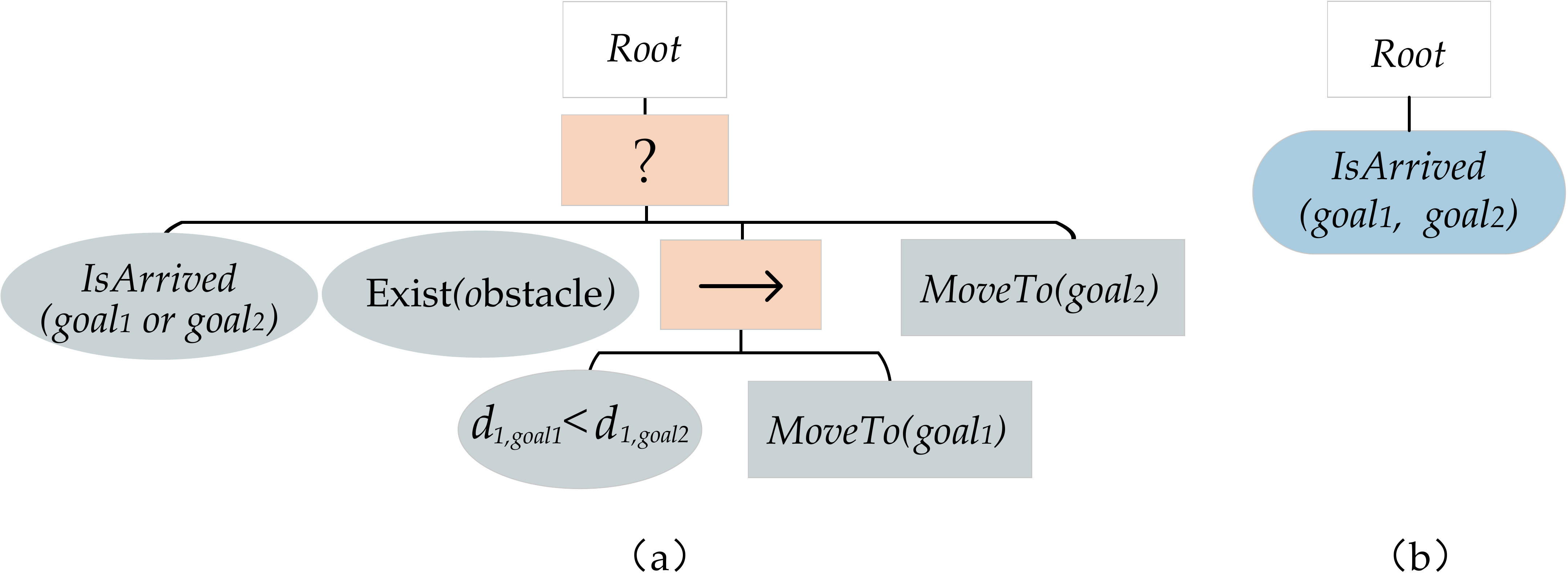}
    \caption{Fig.(a) shows the traditional method for robot \(\mathcal{R}_{1}\) to select the nearest goal while considering other robots' states. Fig.(b) illustrates the interactive inference node, which selects a strategy to minimize free energy. Both figures demonstrate the same functionality.}
    \label{fig4}
\end{figure}

We define the observation sets as 
\[
\mathcal{O}^{1}_{\tau} = \{\mathcal{O}^{r1}_{\tau},\, \mathcal{O}^{r2}_{\tau},\, \mathcal{O}^{\text{result}}_{\tau}\}, % <<< changed: result -> \text{result}
\]
where $\mathcal{O}^{r1}_{\tau}$ is robot $\mathcal{R}_{1}$'s position observation, $\mathcal{O}^{r2}_{\tau}$ contains observations from robot $\mathcal{R}_{2}$, and $\mathcal{O}^{\text{result}}_{\tau}$ indicates whether both robots reached their goals simultaneously. % <<< unified O notation
The likelihood matrix is defined as
\[
\mathcal{A}^{1} = \{\mathcal{A}^{1}_{r1},\, \mathcal{A}^{1}_{r2},\, \mathcal{A}^{1}_{\text{result}}\}. % <<< unified subscript result
\]
The matrix $\mathcal{A}^{1}_{r1}$ illustrates the relationship between robot $\mathcal{R}_{1}$'s observable position and its hidden state, with a probability of accurately determining its position at $0.9952$. 
Formally, $\mathcal{A}^{1}_{r1}\{p_{i}, p_{i},:\} = 0.9952$ for $i \in [0,48]$, indicating that robot $\mathcal{R}_{1}$ correctly determines its own position with probability $0.9952$. 
The matrix $\mathcal{A}^{1}_{r2}\{p_{i}, p_{i},:\} = 0.904$ for $i \in [0,48]$ reflects the accuracy of estimating the other robot’s position. % <<< clarified semantics, no deletion
Let $\mathcal{A}^{1}_{\text{result}}$ denote the joint inference result. Success occurs when robots $\mathcal{R}_{1}$ and $\mathcal{R}_{2}$ achieve different goals simultaneously, represented by $\mathcal{A}^{1}_{\text{result}}(1,  p_{24}, p_{26}) = 1$, or vice versa. 
If the robots fail to achieve their goals simultaneously or select the same goal, it results in a task failure, indicated by $\mathcal{A}^{1}_{\text{result}}(2,p_{24}, p_{26}) = 1$, or vice versa. 
If neither robot reaches a goal, the task result is \textit{null}, denoted as $\mathcal{A}^{1}_{\text{result}}(0, :, :)$. % <<< unified notation and style

The transition matrix $\mathcal{B}^{1}$ describes how hidden states evolve over time $\tau$ based on control actions $a_{\tau} \in U$. 
A sequence of control actions is represented as $\pi^{i} = \{ a_{\tau=1}, a_{\tau=2}, \dots, a_{\tau=n} \}$, with $\pi^{i} \in \Pi^{i}$, where $\Pi^{i}$ includes all strategies for robot $\mathcal{R}_{i}$. 
The matrix \( \mathcal{B}^{1} = \{ \mathcal{B}^{1}_{r1},\, \mathcal{B}^{1}_{r2} \} \) % <<< changed: added r1, r2 subscript for clarity
consists of the state transition matrices for robot \( \mathcal{R}_{1} \) and robot \( \mathcal{R}_{2} \). 
Set $\mathcal{B}^{i}\{ p_{\text{next}}, p_{\text{cur}}, a_{\tau} \} = 1.0$, where \( a_{\tau} \) is the robot's action at time \( \tau \), \( p_{\text{cur}} \) is its current position, and \( p_{\text{next}} \) is the position after action \( a_{\tau} \). % <<< unified next/cur notation

\begin{table}[H]
\label{tab:pref}
\centering
\scriptsize
\caption{Setting Strategy Priorities in Experiments}
\begin{tabular}{ccc}
\hline
Planning         & Precondition  & Postcondition        \\ \hline
$\pi^{i}_{obs}$      & \textit{Exist($obstacle$)}       & $l^{i}_{obs}=-(\max(\mathcal{C}^{i}_{ri})+1)$ \\ % <<< corrected math syntax
$\pi^{i}_{points}$ & \begin{tabular}[c]{@{}c@{}} \textit{!Exist($obstacle$)}\\  \textbf{And} \textit{!IsArrived(points)}\end{tabular} &$l^{i}_{add}=\max(\mathcal{C}^{i}_{ri})+1$  \\ 
$\pi^{i}_{goal}$     & \begin{tabular}[c]{@{}c@{}}\textit{!Exist($obstacle$)}\\ \textbf{And} \textit{!IsArrived($goal$)}\\ \textbf{And} \textit{IsArrived($points$)}\end{tabular}   & $l^{i}_{goal}=\max(\mathcal{C}^{i}_{ri})+1$           \\ 
$\pi^{i}_{stop}$ &\textit{IsArrived($goal$)} &-\\\hline
\end{tabular}
\end{table}

The preference matrix \(\mathcal{C}^{i}\), aligned with the observation matrix $\mathcal{O}^{i}_{\tau}$, indicates preferences for goal locations regarding task outcomes, as defined in
\[
\mathcal{C}^{1} = \{ \mathcal{C}^{1}_{r1},\, \mathcal{C}^{1}_{r2},\, \mathcal{C}^{1}_{\text{result}} \}, % <<< changed: unified \text{result}
\]
where $\mathcal{C}^{1}_{\{r1,r2\}}\{p_{24},p_{26}\}=1$, and  $\mathcal{C}^{1}_{\text{result}}\{\text{success}\}=1$. These are set based on the Pref-Weights in Table~\ref{tab:pref}. % <<< clarified text labels

We derive the probability distributions of strategies in \( \Pi^{1} \times \Pi^{2} \) based on robot configurations. 
The robots select and execute the strategies with the highest probabilities, resulting in the following distributions:
\[
\begin{aligned}
\mathcal{P}_{g1-g1} &= [0.151\times10^{-7}, 0.566\times10^{-6}, 0.995\times10^{-7}, 0.210\times10^{-6}],\\
\mathcal{P}_{g1-g2} &= [0.405\times10^{-7}, 0.357\times10^{-7}, 0.081, 0.072],\\
\mathcal{P}_{g2-g1} &= [0.746\times10^{-7}, 0.847, 0.682\times10^{-7}, 0.573\times10^{-6}],\\
\mathcal{P}_{g2-g2} &= [0.133\times10^{-7}, 0.117\times10^{-7}, 0.408\times10^{-7}, 0.360\times10^{-7}].
\end{aligned}
\]
Here, \( \mathcal{P}_{g_{i}-g_{j}} \) represents the probability distribution of robot \( \mathcal{R}_{1} \) moving toward \( goal_{i} \) while robot \( \mathcal{R}_{2} \) moves toward \( goal_{j} \). 
Robot \( \mathcal{R}_{1} \) will follow the path \(\{ p_{28}, p_{29}, p_{30}, p_{31}, p_{32}, p_{33}, p_{26} \}\), while robot \( \mathcal{R}_{2}\) will follow the path \(\{ p_{2}, p_{1}, p_{8}, p_{15}, p_{22}, p_{23}, p_{24} \}\).\par

\subsection{Interactive Inference Nodes for Conflict Handling}

We demonstrate how robots resolve task conflicts in a dynamic environment, aiming to meet the expected postcondition matrix \( \pi^{1}_{\text{goal}}.\text{postc} \) for achieving the strategy \( \pi^{1}_{\text{stop}} \) (see Table~\ref{tab:pref}). % <<< added \text{goal} for consistency

\begin{figure}[h]
    \centering
    \includegraphics[width=0.6\textwidth]{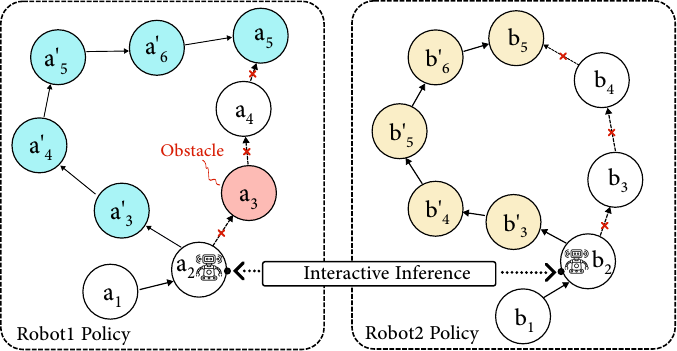}
    \caption{Robot $\mathcal{R}_{1}$, located at position $a_{2}$, detects an obstacle at position $a_{3}$. Consequently, robot $\mathcal{R}_{1}$ abandons its original path $\{ a_{1}, a_{2}, a_{3}, a_{4}, a_{5} \}$ and modifies it to $\left \{a_{2},a'_{3},a'_{4},a'_{5},a'_{6},a_{5} \right \}$. To ensure that it reaches the destination $b_{5}$ simultaneously with robot $\mathcal{R}_{1}$, robot $\mathcal{R}_{2}$, positioned at \(b_{2}\), engages in interactive inference with robot $\mathcal{R}_{1}$ and acquires a new path $\left \{ b_{2},b'_{3},b'_{4},b'_{5},b'_{6},b_{5} \right \}$.}
    \label{fig5}
\end{figure}

While pursuing their goals, the robots evaluate execution conditions in real time at the interactive inference nodes (Algorithm~1, Line~17). 
When executing \( \pi^{1}_{\text{goal}} \), robots must ensure no obstacles block the current path, a precondition \( \pi^{1}_{\text{goal}}.\text{prec} = \{\mathcal{E}^{1}_{r1}, \mathcal{E}^{1}_{r2}, \mathcal{E}^{1}_{\text{result}}\} \). % <<< unified subscript
In the current strategy, \( \mathcal{E}^{1}_{r1}\{p_{29}, p_{30}, p_{31}, p_{32}, p_{33}, p_{26}\} = 0 \) and \( \mathcal{E}^{1}_{r2}\{p_{1}, p_{8}, p_{15}, p_{22}, p_{23}, p_{24}\} = 0 \). 
If a temporary obstacle is added at \( p_{30} \) on robot \( \mathcal{R}_{1} \)'s path, the observation logical quantity for robot \( \mathcal{R}_{1} \) is \( \mathcal{L}^{1} = \{l^{1}_{r1}, l^{1}_{r2}, l^{1}_{\text{result}}\} \), % <<< unified result
where \( l^{1}_{r1}\{ p_{30} \} = l^{1}_{obs}=-1 \). 
Thus, \( \mathcal{L}^{1} \notin \pi^{1}_{\text{goal}}.\text{prec} \). 
The environmental logical variable is added to the preference matrix \( \mathcal{C}^{1} \), resulting in \( \mathcal{C}^{1}_{r1}\{ p_{30} \} = -1 \), \( \mathcal{C}^{1}_{r1}\{ p_{24}, p_{26} \} = 1 \), and \( \mathcal{C}^{1}_{r2}\{ p_{24}, p_{26}\} = 1 \). % <<< unified indices

At this stage, robot \( \mathcal{R}_{1} \) generates several obstacle avoidance strategies, denoted as \(\Pi^{1}_{\text{obs}}\). These strategies are transformed into the path sets \( \{p_{29}, p_{22}, p_{15}, p_{16}, p_{17}, p_{24}\} \) and \( \{p_{29}, p_{36}, p_{37}, p_{38}, p_{31}, p_{32}, p_{33}, p_{26}\} \). \par

Consequently, the strategy set for robot \(\mathcal{R}_{1}\) becomes \(\Pi^{1}=\Pi^{1}_{\text{obs}} \cup \pi^{1}_{\text{goal}}\). Meanwhile, robot \(\mathcal{R}_{2}\), located at \(p_{8}\), generates a new strategy set $\Pi^{2}_{\text{goal}}$, which includes \(\{p_{1},p_{8},p_{9},p_{10},p_{17},p_{24}\}\) and \(\{p_{1},p_{8},p_{9},p_{10},p_{11},p_{18},p_{19},p_{26}\}\). Robot \(\mathcal{R}_{2}\) incorporates its current strategy into \(\Pi^{2}=\Pi^{2}_{\text{goal}} \cup \pi^{2}_{\text{goal}}\). After combining the two strategy sets and executing interactive inference, the resulting strategy distribution is $\mathcal{P}(\Pi^{1} \times \Pi^{2}) = [0.784\times10^{-6}, 0.166\times10^{-6}, 0.521\times10^{-7}, 0.999, 0.156\times10^{-6}, 0.134\times10^{-7}, 0.898\times10^{-13}$, $0.256\times10^{-7}, 0.554\times10^{-14}].$
Based on the principle of minimizing free energy, robot \( \mathcal{R}_{1} \) will move along the path \( \{p_{29}, p_{36}, p_{37}, p_{38}, p_{31}, p_{32}, p_{33}\} \), while robot \( \mathcal{R}_{2} \) will choose the path \( \{p_{1}, p_{8}, p_{9}, p_{10}, p_{17}, p_{24}\} \). Due to the introduction of obstacles, both robots abandon their original strategies in favor of new ones that minimize free energy.\par

\section{Experiments}
\label{sec:experiments}

\subsection{Cooperative Navigation without Predefined Goals}

\begin{figure}[h]
    \centering
    \includegraphics[width=0.5\textwidth]{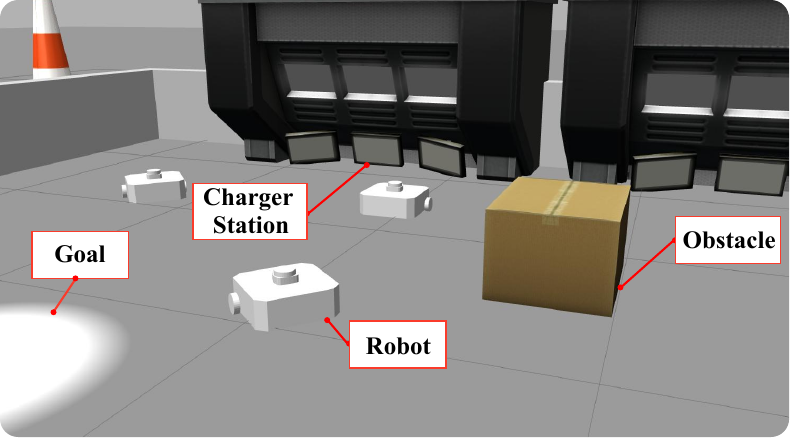}
    \caption{The factory scene features three robots, movable obstacles, and charging stations, with the robots collaborating to reach designated goals.}
    \label{fig7}
\end{figure}

\begin{figure}[h]
    \centering
    \includegraphics[width=0.7\textwidth]{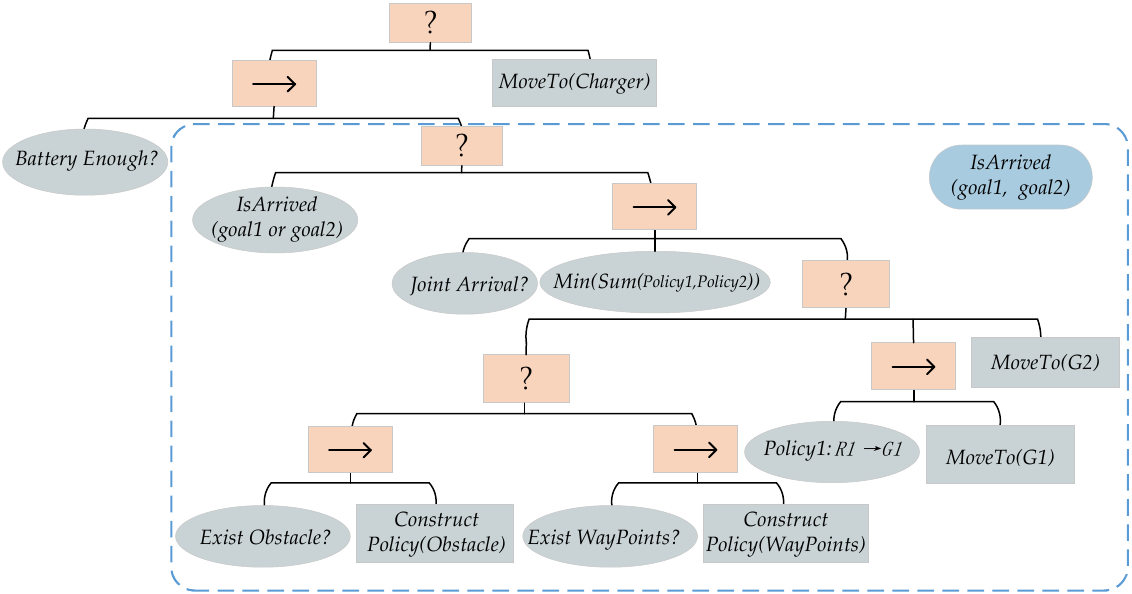}
    \caption{The blue area shows the replacement of traditional control frameworks with interactive inference nodes, emphasizing their compression effects relative to traditional BT nodes.}
    \label{fig8}
\end{figure}

We previously demonstrated how interactive inference nodes manage tasks and conflicts. In this subsection, we use these nodes to create a complex BT for mobile robots performing collaborative tasks in dynamic environments. In our simulation\footnote{https://youtu.be/KX\_oT3IDTf4}, three robots autonomously select goals and adapt to changes, coordinating their movements through a BT controlled by interactive nodes. The task settings for the robot group are shown in Table~I. Fig.~\ref{fig8} shows the independent BT control strategies for each robot. Traditional BTs require 21 nodes, while our approach uses only 5, achieving a 76.2\% reduction in design complexity.\par

\begin{figure}[h]
    \centering
    \includegraphics[width=0.7\textwidth]{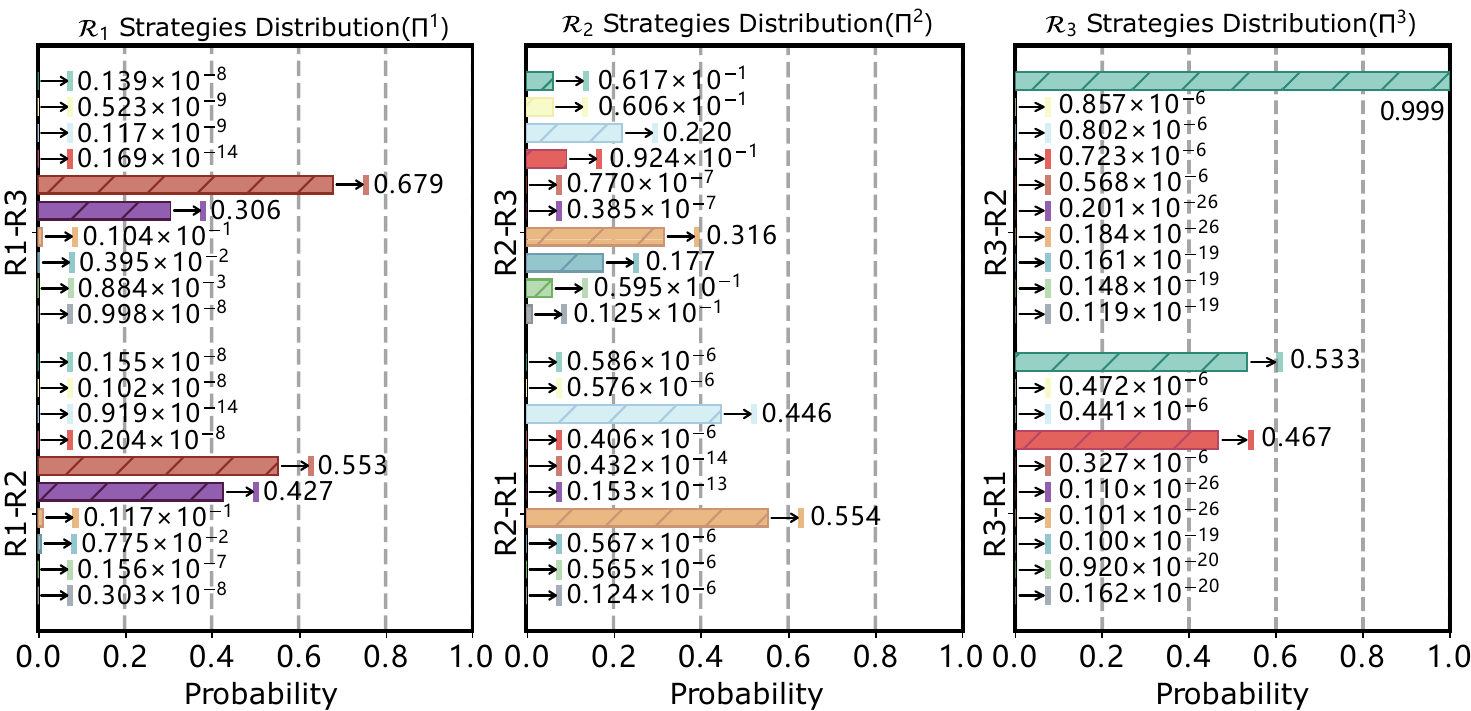}
    \caption{The probability distributions for robots $\mathcal{R}_{1}$, $\mathcal{R}_{2}$, and $\mathcal{R}_{3}$ in the left, middle, and right figures, respectively.}
    \label{fig9}
\end{figure}

Robots \( \mathcal{R}_{1} \), \( \mathcal{R}_{2} \), and \( \mathcal{R}_{3} \) start at positions \( p_{2} \), \( p_{28} \), and \( p_{45} \), respectively. The $goal_{1}$, $goal_{2}$, and $goal_{3}$ are located at \( p_{24} \), \( p_{26} \), and \( p_{32} \), while their charging stations are at \( p_{13} \), \( p_{27} \), and \( p_{41} \). In subsequent experiments, strategy sets \(\Pi^{1}\), \(\Pi^{2}\), and \(\Pi^{3}\) employ the A* algorithm to design paths for each goal, incorporating \( a_{\tau}=\textit{wait} \) into the action strategies for each robot. To reduce computational load, we utilize pairwise interactions—such as between \( \mathcal{R}_{1} \) and \( \mathcal{R}_{2} \), and between \( \mathcal{R}_{1} \) and \( \mathcal{R}_{3} \)—as illustrated in Fig.~\ref{fig9}. % <<< unified notation, clarified text

For instance, \( \mathcal{R}_{1} \) interacts with both \( \mathcal{R}_{2} \) and \( \mathcal{R}_{3} \). Robot \( \mathcal{R}_{1} \) generates strategies for \( goal_{1} \), \( goal_{2} \), or \( goal_{3} \) (Algorithm~1, Line~7). Robots \( \mathcal{R}_{1} \) and \( \mathcal{R}_{2} \) then apply the interactive inference procedure (Algorithm~1, Line~10) using the defined likelihood matrix \( \mathcal{A}^{1} \) and transition matrix \( \mathcal{B}^{1} \). Robot \( \mathcal{R}_{1} \) selects the strategy with the highest posterior probability. The same inference process is repeated between robots \( \mathcal{R}_{1} \) and \( \mathcal{R}_{3} \). In the preference matrix \( \mathcal{C}^{1} \), we set \( \mathcal{C}^{1}_{\{r1, r2, r3\}}\{p_{24}, p_{26}, p_{32}\} = 1 \) and \( \mathcal{C}^{1}_{\text{result}}\{\text{success}\} = 1 \). % <<< unified \text{result}

The sixth strategy has the highest posterior probability: $\mathcal{P}(\pi^{1}_{6}) = 0.553$, $\mathcal{P}(\pi^{2}_{6}) = 0.679$. % <<< changed subscript order
Robot \( \mathcal{R}_{1} \) follows the path \( \{ p_{28}, p_{21}, p_{22}, p_{23}, p_{24}, p_{25}, p_{26} \} \) and communicates this policy intention to robot \( \mathcal{R}_{2} \). 
Robot \( \mathcal{R}_{2} \) then infers its strategy with probabilities of $0.554$ and $0.316$, selecting \( \{ p_{2}, p_{9}, p_{16}, p_{16}, p_{23}, p_{23}, p_{24} \} \) to relay to robot \( \mathcal{R}_{3} \). 
Robot \( \mathcal{R}_{3} \) performs inference based on the intentions of robots \( \mathcal{R}_{1} \) and \( \mathcal{R}_{2} \), selecting \( \{ p_{45}, p_{38}, p_{31}, p_{31}, p_{31}, p_{31}, p_{32} \} \). 
This demonstrates distributed interactive inference among the robots.\par % <<< wording refinement

This section details the task execution process for robot \( \mathcal{R}_{1} \). 
While moving to position \( p_{21} \), it encounters an obstacle at \( p_{23} \). 
The robot updates its environmental observation \( \mathcal{O}^{r1}_{\tau} \) at each cycle, generating logical variables 
\[
\mathcal{L}^{1} = \{ l^{1}_{r1}, l^{1}_{r2}, l^{1}_{r3}, l^{1}_{\text{result}} \}. % <<< unified \text{result}
\]
Robot \( \mathcal{R}_{1} \) assigns a preference value of $-1$ to the obstacle at \( p_{23} \), resulting in a conflict between the obstacle-avoidance and current strategies, as \( \mathcal{L}^{1}\{l^{1}_{r1}\} \notin \textit{!Exist(obstacle)} \) (Algorithm~1, Line~17). 
Robot \( \mathcal{R}_{1} \) adds \( \mathcal{L}^{1}\{l^{1}_{r1}\} \) to matrix \( \mathcal{C}^{1} \) (Algorithm~1, Line~18) and guides robots \( \mathcal{R}_{1} \), \( \mathcal{R}_{2} \), and \( \mathcal{R}_{3} \) to develop new strategies (Algorithm~1, Line~19). 
The preference matrix updates to \( \mathcal{C}^{1}_{r1} = \mathcal{C}^{1}_{r1} + l^{1}_{r1} \). 
Robot \( \mathcal{R}_{1} \) then performs inference (Algorithm~1, Line~21), yielding maximum strategy probabilities of $0.166$ and $0.252$, leading to the path \( \{ p_{21}, p_{14}, p_{15}, p_{16}, p_{17}, p_{24} \} \). 
Robot \( \mathcal{R}_{2} \) selects \( \{ p_{9}, p_{10}, p_{17}, p_{24}, p_{31}, p_{32} \} \), and robot \( \mathcal{R}_{3} \) chooses \( \{ p_{38}, p_{38}, p_{31}, p_{24}, p_{25}, p_{26} \} \). 
An obstacle at \( p_{23} \) causes robot \( \mathcal{R}_{1} \) to switch from $goal_{2}$ to $goal_{1}$, prompting robots \( \mathcal{R}_{2} \) and \( \mathcal{R}_{3} \) to modify their goals and strategies.\par

\begin{figure}[h]
    \centering
    \includegraphics[width=0.7\textwidth]{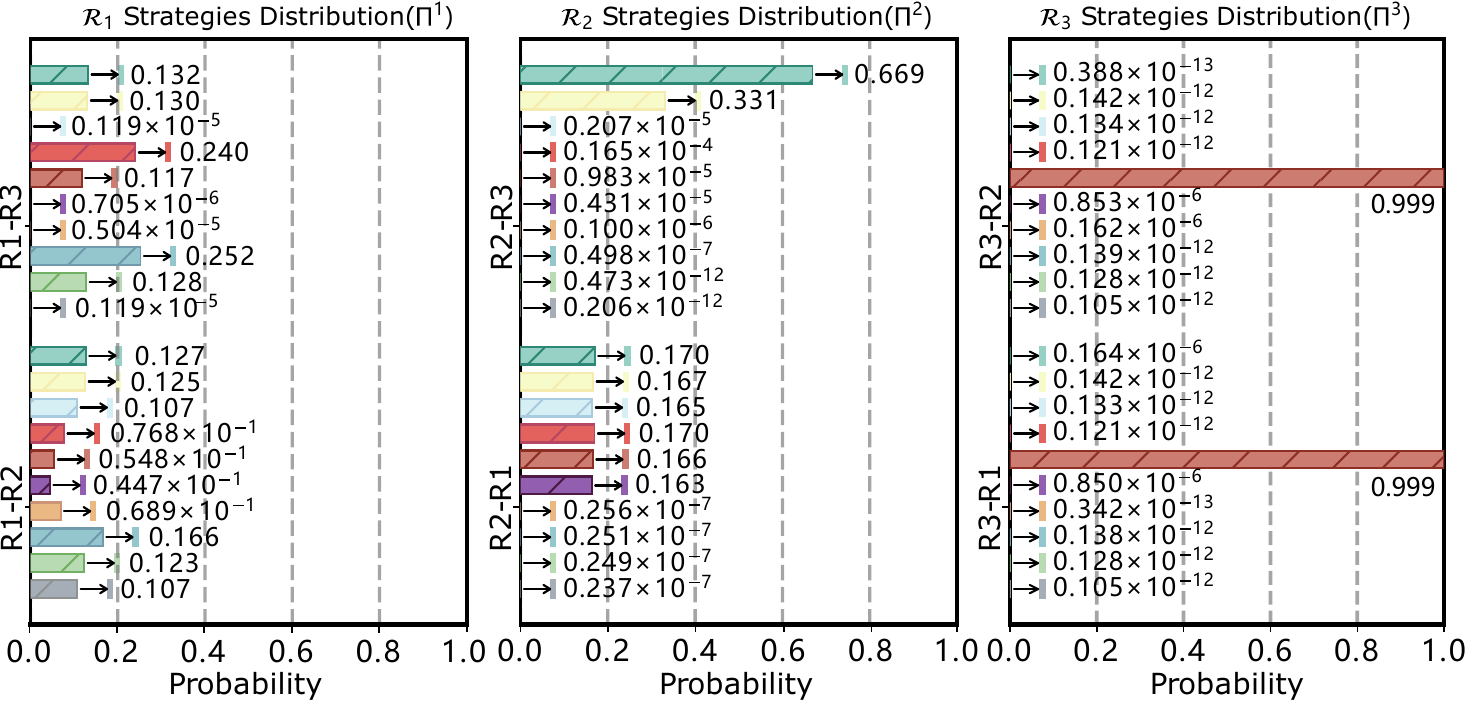}
    \caption{After introducing obstacles into the planned path of robot $\mathcal{R}_{1}$, the distribution of planning strategies for robots $\mathcal{R}_{1}$, $\mathcal{R}_{2}$, and $\mathcal{R}_{3}$.}
    \label{fig10}
\end{figure}

When robot \( \mathcal{R}_{1} \) moves to \( p_{14} \), a new obstacle at \( p_{31} \) blocks robot \( \mathcal{R}_{3} \), while the original obstacle remains. 
The preference matrix for robot \( \mathcal{R}_{3} \) updates to \( \mathcal{C}^{1}_{r3} = \mathcal{C}^{1}_{r3} + l^{1}_{r3} \). % <<< fixed wrong subscript
The updated strategies are \( \{ p_{14}, p_{15}, p_{16}, p_{17}, p_{24}, p_{25}, p_{26} \} \) for robot \( \mathcal{R}_{1} \), \( \{ p_{16}, p_{17}, p_{18}, p_{18}, p_{25}, p_{25}, p_{32} \} \) for robot \( \mathcal{R}_{2} \), and \( \{ p_{38}, p_{39}, p_{32}, p_{25}, p_{18}, p_{17}, p_{24} \} \) for robot \( \mathcal{R}_{3} \).\par

\begin{figure}[h]
    \centering
    \includegraphics[width=0.7\textwidth]{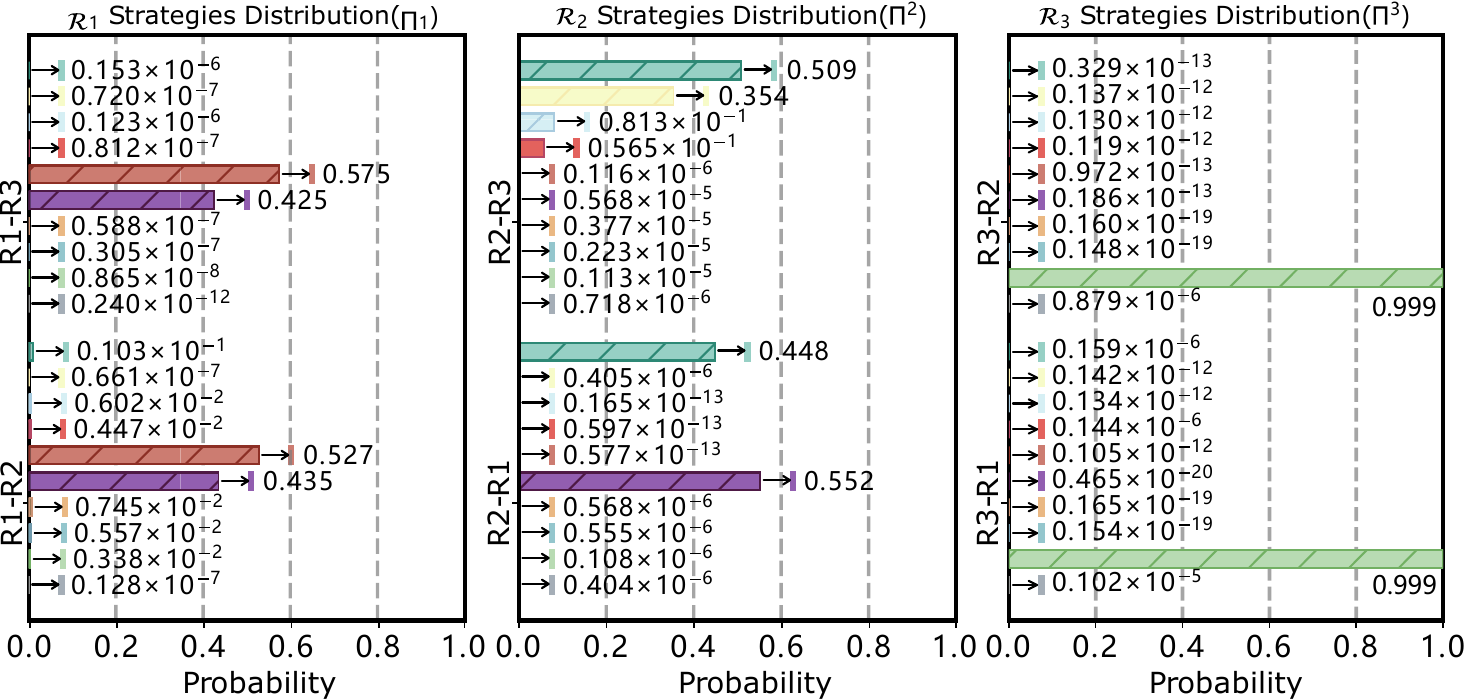}
    \caption{After introducing obstacles into the planned path of robot $\mathcal{R}_{3}$, the distribution of planning strategies for robots $\mathcal{R}_{1}$, $\mathcal{R}_{2}$, and $\mathcal{R}_{3}$.}
    \label{fig11}
\end{figure}

During task execution, we introduce temporary waypoints for robots $\mathcal{R}_{1}$ and \( \mathcal{R}_{3} \) (see Table~I). 
These waypoints take precedence over $\pi^{i}_{\text{goal}}$. % <<< unified goal notation
When robot \( \mathcal{R}_{1} \) moves to \( p_{17} \), robots \( \mathcal{R}_{2} \) and \( \mathcal{R}_{3} \) will move to \( p_{25} \) and \( p_{39} \), respectively. 
Robots \(\mathcal{R}_{1}\) and \(\mathcal{R}_{2}\) add waypoints \(p_{4}\) and \(p_{35}\), with \(l^{1}_{r1}(p_{4}) = 2\) and \(l^{1}_{r3}(p_{35}) = 2\). 
The preference matrices are updated as \( \mathcal{C}^{1}_{r1}=\mathcal{C}^{1}_{r1}+l^{1}_{r1} \) and \( \mathcal{C}^{1}_{r3}=\mathcal{C}^{1}_{r3}+l^{1}_{r3} \). 
To minimize free energy and reach their goals simultaneously, robot \( \mathcal{R}_{1} \) selects the path \( \{ p_{17}, p_{10}, p_{3}, p_{4}, p_{4}, p_{3}, p_{10}, p_{17}, p_{24} \} \), 
robot \( \mathcal{R}_{2} \) chooses \( \{ p_{18}, p_{25}, p_{25}, p_{25}, p_{25}, p_{25}, p_{25}, p_{25}, p_{32} \} \), 
and robot \( \mathcal{R}_{3} \) adopts \( \{ p_{25}, p_{32}, p_{39}, p_{39}, p_{32}, p_{25}, p_{25}, p_{25}, p_{26} \} \).\par

\begin{figure}[h]
    \centering
    \includegraphics[width=0.7\textwidth]{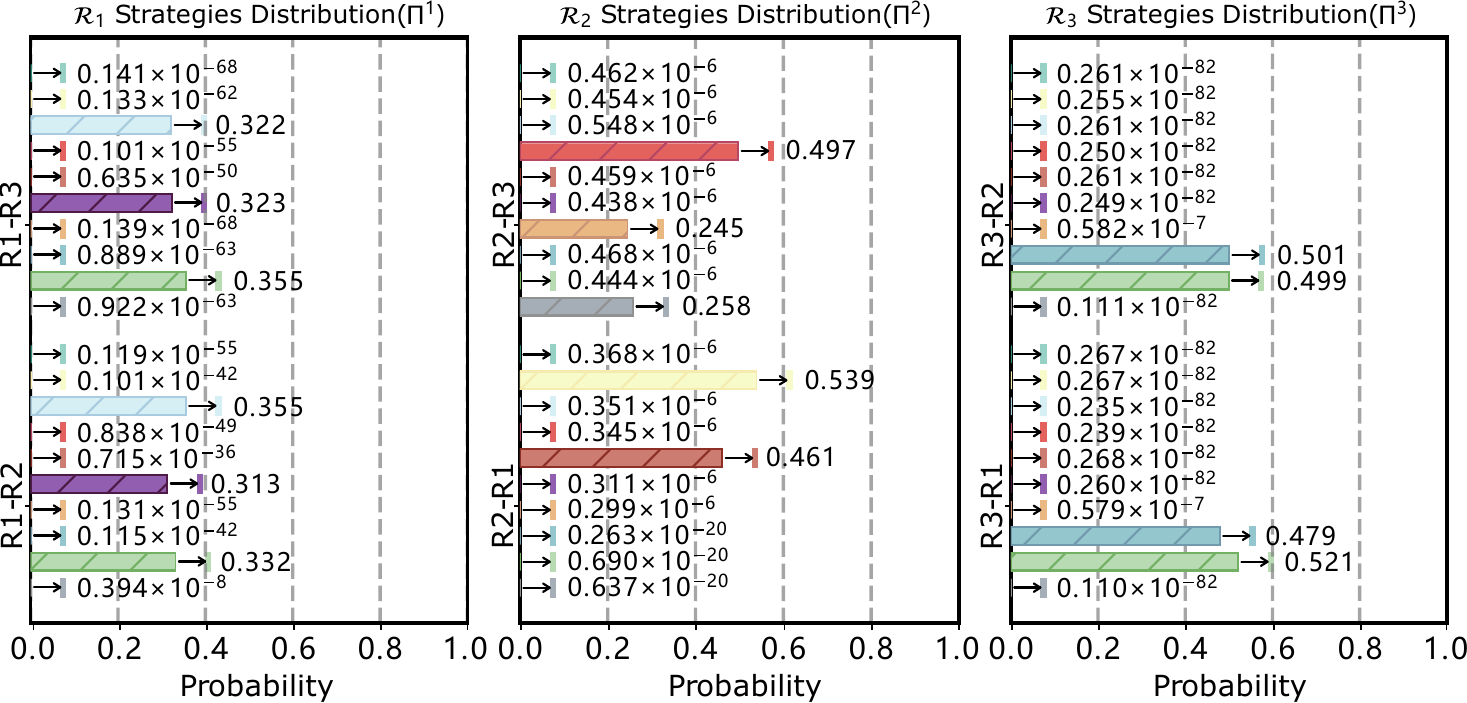}
    \caption{After adding waypoints for robots $\mathcal{R}_{1}$ and $\mathcal{R}_{3}$, the distribution of planning strategies for robots $\mathcal{R}_{1}$, $\mathcal{R}_{2}$, and $\mathcal{R}_{3}$.}
    \label{fig12}
\end{figure}

Distributed BTs allow interactive BT nodes to cooperate with traditional BT nodes, ensuring robot coordination and autonomous decision-making. 
Each robot has a \textit{Battery Enough?} condition node and a \textit{MoveToCharger} action node, enabling them to autonomously detach from the swarm and reach a charging station when battery levels are low. 
All strategies must meet the \textit{Battery Enough} precondition. 
When a robot runs the interactive node \textit{MoveToGoal}, it returns a \textit{running} status each cycle. 
If the battery is low, the \textit{MoveToGoal} node returns a \textit{failure} status, causing disconnection. 
When the \textit{Battery Enough?} node shows a \textit{failure} status, the robot activates the \textit{MoveToCharger} node. 
For instance, robot \( \mathcal{R}_{1} \) moves autonomously from position \( p_{4} \) to Charging Station~1 (\( p_{13} \)), disconnecting from robots \( \mathcal{R}_{2} \) and \( \mathcal{R}_{3} \). 
Robots \( \mathcal{R}_{2} \) and \( \mathcal{R}_{3} \) wait for \( \mathcal{R}_{1} \) to reconnect after recharging. 
Once charged, \( \mathcal{R}_{1} \) returns to its position and resumes its strategy. 
If robot \( \mathcal{R}_{3} \) detects a low battery at \( p_{25} \), it goes to Charging Station~3 (\( p_{41} \)) while \( \mathcal{R}_{1} \) and \( \mathcal{R}_{2} \) wait. 
Upon reunion, all three reach their destination together, adjusting their strategies to \( \pi^{\{1,2,3\}} = \pi_{\text{stop}} \). % <<< unified stop notation
This experiment shows that robots can effectively use multi-robot interactive nodes with traditional BT nodes, maintaining coordinated movement and autonomous disengagement during group tasks.\par

\begin{table}[H]
\centering
\scriptsize
\caption{QUANTITATIVE COMPARISON OF BEHAVIOR TREE COMPLEXITY}
\label{tab:bt_complexity_comparison}
\begin{tabular}{lcccc}
\hline
 & \textbf{Traditional} & \textbf{IIBT-Node} & \textbf{Absolute} & \textbf{Relative} \\
\textbf{Component} & \textbf{BT} & \textbf{Approach} & \textbf{Reduction} & \textbf{Reduction} \\
\hline
Sequence Nodes  & 5 & 1 & 4 & 80\% \\
Fallback Nodes  & 4 & 1 & 3 & 75\% \\
Condition Nodes & 7 & 1 & 6 & 85\% \\
Action Nodes    & 5 & 2 & 3 & 60\% \\
\hline
\textbf{Total Nodes} & \textbf{21} & \textbf{5} & \textbf{16} & \textbf{76.2\%} \\
\hline
Tree Depth & 8 & 2 & 6 & 75\% \\
Design Complexity & High & Low & - & - \\
\hline
\end{tabular}
\end{table}

As illustrated in Fig.~\ref{fig8}, the proposed IIBT architecture replaces large portions of traditional BT control logic (shown in blue) with compact interactive inference nodes. 
These nodes integrate action selection, precondition checking, and inter-robot coordination within a single probabilistic framework, significantly simplifying the tree topology. 
While a conventional BT relies on multiple layers of Sequence, Fallback, and Condition nodes to handle task transitions and recovery behaviors, the IIBT node internally performs these functions through belief updates and free-energy minimization.

To quantitatively evaluate this structural compression, Table~\ref{tab:bt_complexity_comparison} compares the node composition and tree depth of the traditional BT against the proposed IIBT approach. 
The results confirm that embedding inference capabilities into BT nodes yields a 76.2\% reduction in node count and a 75\% decrease in tree depth, effectively transforming a deep, rule-based hierarchy into a compact, adaptive decision structure.

\subsection{Multirobot Cooperative Object Placement Real-World Experiment} % <<< changed: title capitalization
Building upon the navigation experiments presented in the previous subsection, this study extends the proposed Interactive Inference Behavior Tree (IIBT) framework from multi-robot motion coordination to a real-world collaborative object placement task. 
The goal of this experiment is to quantitatively evaluate the generality and robustness of the IIBT-Node architecture under physical conditions with sensory noise and actuation uncertainty. 
Specifically, the experiment aims to verify (1) the practical deployability of the BT-based control framework on real robot platforms, and (2) the robustness of the interactive inference mechanism when faced with perceptual and control disturbances.

Within this setup, the robot agents are capable of sharing real-time task states and executing coordinated movements. 
The action space for each robot is defined as 
\[
U = \{\text{moveTo(goal)},~\text{pick(obj)},~\text{place(obj)},~\text{idle}\}. % <<< changed
\]
To encompass the full range of potential cooperative behaviors, a joint strategy set $\Pi^{\{1,2\}}$ is constructed as
\[
\Pi^{\{1,2\}} = \{\pi_{0}, \pi_{1}, \dots, \pi_{15}\}, % <<< changed
\]
where each joint policy $\pi_i$ is a tuple $\pi_i = (a_{\mathcal{R}_1}, a_{\mathcal{R}_2})$, with $a_{\mathcal{R}_j} \in U$. 
Accordingly, the complete set of joint policy combinations is enumerated as:
{\scriptsize
\setlength{\arraycolsep}{2pt}
\begin{align*}
\pi_{0}  &= (\text{moveTo(goal)},~\text{moveTo(goal)}), &
\pi_{1}  &= (\text{moveTo(goal)},~\text{pick(obj)}), \\
\pi_{2}  &= (\text{moveTo(goal)},~\text{place(obj)}), &
\pi_{3}  &= (\text{moveTo(goal)},~\text{idle}), \\
\pi_{4}  &= (\text{pick(obj)},~\text{moveTo(goal)}), &
\pi_{5}  &= (\text{pick(obj)},~\text{pick(obj)}), \\
\pi_{6}  &= (\text{pick(obj)},~\text{place(obj)}), &
\pi_{7}  &= (\text{pick(obj)},~\text{idle}), \\
\pi_{8}  &= (\text{place(obj)},~\text{moveTo(goal)}), &
\pi_{9}  &= (\text{place(obj)},~\text{pick(obj)}), \\
\pi_{10} &= (\text{place(obj)},~\text{place(obj)}), &
\pi_{11} &= (\text{place(obj)},~\text{idle}), \\
\pi_{12} &= (\text{idle},~\text{moveTo(goal)}), &
\pi_{13} &= (\text{idle},~\text{pick(obj)}), \\
\pi_{14} &= (\text{idle},~\text{place(obj)}), &
\pi_{15} &= (\text{idle},~\text{idle}).
\end{align*}
}

This joint strategy space represents all possible combinations of cooperative actions between robots $\mathcal{R}_1$ and $\mathcal{R}_2$, forming the foundation for subsequent inference-based policy selection. 
Within the proposed framework, the update of each robot’s task preference matrix $\mathcal{C}^{i}$ is modulated by its corresponding logical variable set 
\[
\mathcal{L}^{i} = \{l^{i}_{\text{loc}}, l^{i}_{\text{hold}}, l^{i}_{\text{place}}, l^{i}_{\text{free}}\}. % <<< changed
\]
The logical variables $\mathcal{L}^{i}$ serve as symbolic representations of discrete task states and act as logical priors that shape the preference update in $\mathcal{C}^{i}$ according to the robot’s perceived progress and task requirements. 
The definitions of the hidden state vector $s^{i}$ and the corresponding logical variables $\mathcal{L}^{i}$ used in this experiment are listed in Table~\ref{tab:state-definition}.

\begin{table}[H]
\centering
\scriptsize
\caption{Definitions of Robot Hidden States and Logical Variables ($l^{i}_{\cdot} \in \mathcal{L}^{i}$)}
\label{tab:state-definition}
\begin{tabular}{ccc}
\hline
\textbf{Hidden State} & \textbf{Logical Variable} & \textbf{Semantic Description} \\ \hline
$s^{i}_{\text{loc}}$ & $l^{i}_{\text{loc}}$ & Posterior belief of reaching goal location \\
$s^{i}_{\text{hold}}$ & $l^{i}_{\text{hold}}$ & Posterior belief of grasping the object \\
$s^{i}_{\text{place}}$ & $l^{i}_{\text{place}}$ & Posterior belief of object placement \\
$s^{i}_{\text{free}}$ & $l^{i}_{\text{free}}$ & Posterior belief of being idle or task-free \\ \hline
\end{tabular}
\end{table}

The preconditions and postconditions governing the execution of each robotic action are defined in Table~\ref{tab:actions}. 
Each action corresponds to a logical transition that determines whether the associated condition can be executed within the IIBT-Node. 
Upon action completion, the postconditions update the logical variable set $\mathcal{L}^i$ and indirectly modify the task preference matrix $\mathcal{C}^i$ through additive adjustments to the relevant entries.

\begin{table}[H]
\centering
\caption{Action specifications with preconditions and postconditions}
\label{tab:actions}
\begin{tabular}{ccc}
\hline
Action      & Preconditions                                                                              & Postconditions                                     \\ \hline
moveTo(loc) & !IsReached(loc)                                                                          & $l^{i}_{\text{loc}} = \max(\mathcal{C}^{i}) + 1$   \\ % <<< changed
pick(obj)   & \begin{tabular}[c]{@{}c@{}}IsReached(loc)\\ !IsHolding\end{tabular}                      & $l^{i}_{\text{hold}} = \max(\mathcal{C}^{i}) + 1$  \\ % <<< changed
place(obj)  & \begin{tabular}[c]{@{}c@{}}IsReached(loc)\\ IsHolding(obj)\\ !IsPlaced(obj,loc)\end{tabular} & $l^{i}_{\text{place}} = \max(\mathcal{C}^{i}) + 1$ \\ \hline
\end{tabular}
\end{table}

As shown, each postcondition reflects the logical progression of the task: 
for instance, executing \textit{moveTo(loc)} increases the preference for the \textit{IsReached(loc)} state ($l^{i}_{\text{loc}}$), 
while \textit{pick(obj)} and \textit{place(obj)} increment the corresponding $s^i_{\text{hold}}$ and $s^i_{\text{place}}$ beliefs. 
The \textit{idle} action, in contrast, maintains the $s^i_{\text{free}}$ belief, indicating no active task engagement. % <<< changed

To enable probabilistic reasoning within the IIBT-Node, this section formalizes the probabilistic matrices that constitute the core of the inference process. 
A likelihood matrix $\mathcal{A}^{i}$ is defined for each robot $\mathcal{R}_i$, modeling the conditional relationship between the hidden state $s^i$ and the observation matrix $\mathcal{O}^{i}_{\tau}$ as $\mathcal{P}(\mathcal{O}^{i}_{\tau} \mid s^i)$. % <<< changed
This matrix captures how sensory evidence updates the robot’s belief state under perceptual uncertainty.

\begin{equation}
\mathcal{A}^{i} =
\begin{bmatrix}
0.9 & 0.025 & 0.025 & 0.025 \\
0.025 & 0.9 & 0.025 & 0.025 \\
0.025 & 0.025 & 0.9 & 0.025 \\
0.025 & 0.025 & 0.025 & 0.9 \\
\end{bmatrix}. % <<< changed (removed subscript ri)
\end{equation}

Each row of $\mathcal{A}^{i}$ corresponds to a specific hidden state $s^i_{k}$ and represents the conditional probability distribution $\mathcal{P}(\mathcal{O}^i_{\tau} \mid s^i_{k})$ over possible observations. 
For example, the diagonal entry of 0.9 in the first row indicates a 90\% probability of correctly perceiving the intended feature (e.g., target location) when the system is in that state. 
The low off-diagonal probabilities (0.025) represent potential sensor noise or ambiguous observations. 
The observation $\mathcal{O}^{i}_{\tau}$ is thus generated according to this mapping, providing the basis for Bayesian belief updates during inference.

Subsequently, a set of state transition matrices $\mathcal{B}^{i}_{\pi}$ is defined to characterize the probabilistic dynamics of $s^{i}$ under the execution of each action policy $\pi$. 
Each element $\mathcal{B}^{i}_{\pi}(s'|s)$ denotes the probability of transitioning from state $s$ to $s'$ given $\pi$. 
Together with $\mathcal{A}^{i}$, these transition models complete the probabilistic generative structure of the IIBT-Node.

\begin{figure}[ht]
\centering
\includegraphics[width=1\textwidth]{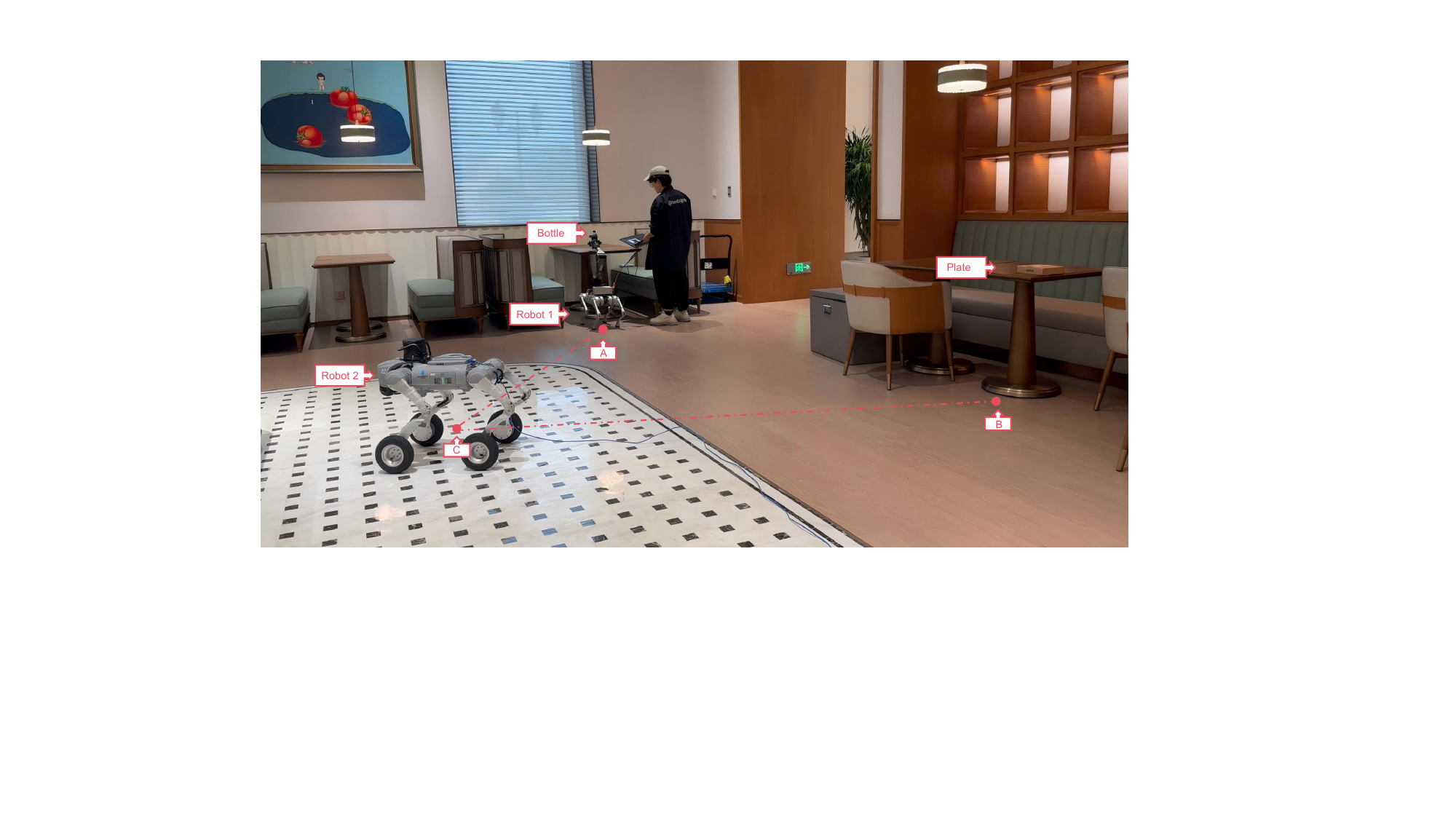}
\caption{Experimental scenario illustrating collaborative object grasping and placement by two quadruped robots.}
\label{fig:real_experiment}
\end{figure}

Having defined the probabilistic model and action semantics of the IIBT-Node, this section introduces the specific joint task designed for real-world multi-robot evaluation. 
As illustrated in Fig.~\ref{fig:real_experiment}, the experimental environment comprises three key locations: Goal A, Goal B, and the Rendezvous Point C. 
This setup extends the previous navigation experiment into a cooperative manipulation domain, enabling evaluation of the proposed framework under more complex physical interactions and goal dependencies.

Two quadruped robots, $\mathcal{R}_1$ and $\mathcal{R}_2$, are deployed to execute a collaborative object placement mission. 
Each robot is equipped with a front-mounted manipulator and onboard cameras for local perception, allowing them to share task-relevant state information in real time. 
Robot $\mathcal{R}_1$ is assigned to navigate toward Goal A, grasp a bottle, and transport it to Rendezvous Point C. 
Robot $\mathcal{R}_2$ is tasked with navigating to Goal B, picking up a plate, and likewise delivering it to the same rendezvous point. 
The shared objective is for $\mathcal{R}_1$ to precisely place the bottle onto the plate held by $\mathcal{R}_2$ at Location C, thereby completing the cooperative manipulation task. 
This configuration allows both agents to infer and update their action policies from the joint policy set $\Pi^{\{1,2\}} = \{\pi_0, \pi_1, ..., \pi_{15}\}$ based on their joint observations $\mathcal{O}^{\{1,2\}}_{\tau}$ and preference matrices $\mathcal{C}^{\{1,2\}}$. % <<< changed

In this cooperative manipulation task, temporal dependencies between actions emerge naturally. 
A critical inter-robot dependency exists: the successful execution of the \textit{place} action by $\mathcal{R}_1$ is contingent upon $\mathcal{R}_2$ having already positioned the plate at Location C. 
If $\mathcal{R}_1$ reaches Location C before $\mathcal{R}_2$ completes its placement action, it must enter a waiting state ($s^{1}_{\text{free}}$) until it detects that the plate is in place. % <<< changed
This dependency requires each robot to reason not only about its own latent state $s^i_{\tau}$ and selected action $a^i_{\tau}$, but also to infer and adapt to the evolving strategy of its partner in real time, thereby encapsulating the core challenge of multi-agent interactive inference.

\begin{figure}[ht]
\centering
\includegraphics[width=0.6\textwidth]{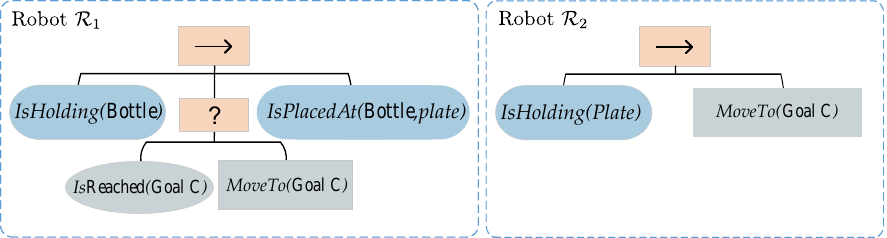}
\caption{Interactive inference behavior trees for (left) robot $\mathcal{R}_1$ and (right) robot $\mathcal{R}_2$.}
\label{fig:bt_structures}
\end{figure}

To evaluate this mechanism, a baseline was implemented using a traditional BT design approach. 
Fig.~\ref{fig:bt_structures} presents the executable BT structures for $\mathcal{R}_1$ and $\mathcal{R}_2$, both incorporating the proposed interactive inference nodes (IIBT-nodes). 
For robot $\mathcal{R}_1$, the IIBT-nodes include \textit{IsHolding(Bottle)} and \textit{IsPlaced(Bottle,Plate)}, complemented by conventional BT nodes such as the condition \textit{IsReached(Goal A)} and the action \textit{moveTo(Goal C)}. % <<< changed
For robot $\mathcal{R}_2$, the IIBT-nodes include \textit{IsHolding(Plate)} and \textit{IsMovingWith(Bottle,Plate)}, together with standard nodes such as the condition \textit{IsReached(Goal C)} and the action \textit{moveTo(Goal C)}. % <<< changed

\begin{figure}[ht]
\centering
\includegraphics[width=0.7\textwidth]{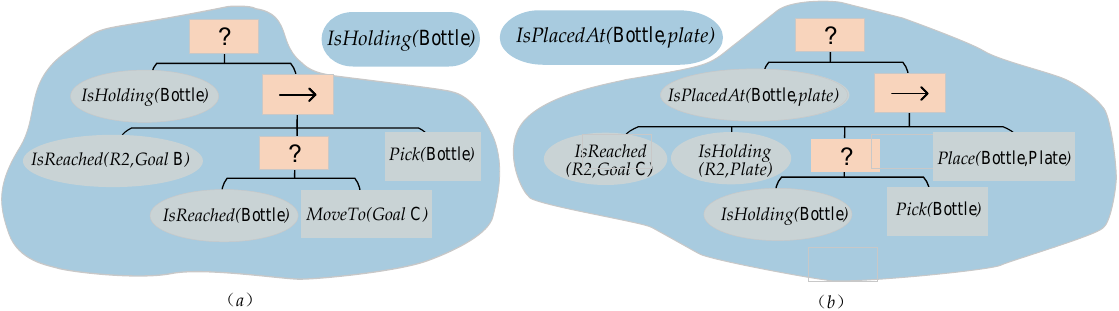}
\caption{Structural compression comparison of BT nodes for Robot $\mathcal{R}_{1}$.}
\label{fig:r1_bt_compression}
\end{figure}

\begin{figure}[ht]
\centering
\includegraphics[width=0.3\textwidth]{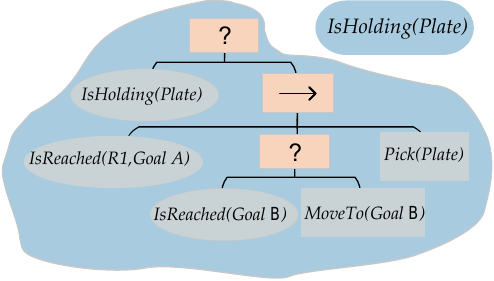}
\caption{Structural compression comparison of BT nodes for Robot $\mathcal{R}_{2}$.}
\label{fig:r2_bt_compression}
\end{figure}

Fig.~\ref{fig:r1_bt_compression} further illustrates the BTs that are functionally equivalent to the aforementioned IIBT-nodes, demonstrating the structural expansion required to achieve the same logical expressiveness in a conventional BT framework. 
Specifically, the equivalent BT for the \textit{IsHolding(Bottle)} node of $\mathcal{R}_1$ is shown in Fig.~\ref{fig:r1_bt_compression}(a), while the one corresponding to \textit{IsPlaced(Bottle,Plate)} is depicted in Fig.~\ref{fig:r1_bt_compression}(b). 
Similarly, for $\mathcal{R}_2$, the equivalent BTs for the \textit{IsHolding(Plate)} and \textit{IsMovingWith(Bottle,Plate)} nodes are presented in Fig.~\ref{fig:r2_bt_compression}(a) and Fig.~\ref{fig:r2_bt_compression}(b), respectively. 
Due to distinct task objectives, the architectures of the two robots’ BTs differ, highlighting the modular adaptability of the proposed IIBT framework.

To quantify the efficacy of our approach, Table~\ref{tab:bt1_complexity_comparison} provides a node count comparison between the interactive inference BT (Fig.~\ref{fig:bt_structures}(a)) and its functionally equivalent traditional counterpart (Fig.~\ref{fig:r1_bt_compression}(a)). The conventional implementation necessitates 4 Sequence nodes, 9 Fallback nodes, 9 Condition nodes, and 11 Action nodes, totaling 33 nodes. In contrast, our approach reduces these requirements to 1 Sequence node, 1 Fallback node, 1 Condition node, and 3 Action nodes, totaling merely 6 nodes—achieving a compression rate of 81.8\%. Furthermore, the structural depth is compressed from 7 layers to 3 layers, substantially alleviating design complexity.

Similarly, Table~\ref{tab:bt2_complexity_comparison} presents comparisons for robot $\mathcal{R}_{1}$'s complete decision-making control model. The traditional BT requires 10 total nodes, compressed to just 6 nodes using our method—a 70\% reduction—with depth reduced from 5 layers to 2 layers.

\begin{table}[!ht]
\centering
\caption{QUANTITATIVE COMPARISON OF $\mathcal{R}_{1}$ BEHAVIOR TREE COMPLEXITY}
\label{tab:bt1_complexity_comparison}
\begin{tabular}{lcccc}
\hline
 & \textbf{Traditional} & \textbf{Ours} & \textbf{Absolute} & \textbf{Relative} \\
\textbf{Component} & \textbf{BT} & \textbf{BT} & \textbf{Reduction} & \textbf{Reduction} \\
\hline
Sequence Nodes & 4 & 1 & 3 & 75\% \\
Fallback Nodes & 9 & 1 & 8 & 88.9\% \\
Condition Nodes & 9 & 1 & 8 & 88.9\% \\
Action Nodes & 11 & 3 & 8 & 73\% \\
\hline
\textbf{Total Nodes} & \textbf{33} & \textbf{6} & \textbf{27} & \textbf{81.8\%} \\
\hline
Tree Depth & 7 & 3 & 4 & 57.1\% \\
Design Complexity & High & Low & - & - \\
\hline
\end{tabular}
\end{table}

\begin{table}[!ht]
\centering
\caption{QUANTITATIVE COMPARISON OF $\mathcal{R}_{2}$ BEHAVIOR TREE COMPLEXITY}
\label{tab:bt2_complexity_comparison}
\begin{tabular}{lcccc}
\hline
 & \textbf{Traditional} & \textbf{Ours} & \textbf{Absolute} & \textbf{Relative} \\
\textbf{Component} & \textbf{BT} & \textbf{BT} & \textbf{Reduction} & \textbf{Reduction} \\
\hline
Sequence Nodes & 2 & 1 & 1 & 50\% \\
Fallback Nodes & 2 & 0 & 2 & 100\% \\
Condition Nodes & 3 & 0 & 3 & 100\% \\
Action Nodes & 3 & 2 & 1 & 33.3\% \\
\hline
\textbf{Total Nodes} & \textbf{10} & \textbf{3} & \textbf{7} & \textbf{70\%} \\
\hline
Tree Depth & 5 & 2 & 3 & 60\% \\
Design Complexity & High & Low & - & - \\
\hline
\end{tabular}
\end{table}

At the initial time step ($\tau = 0$), both robots $\mathcal{R}_1$ and $\mathcal{R}_2$ are in the \textit{idle} state. 
Their joint observation matrix $\mathcal{O}^{\{1,2\}}_{\tau=0}$, composed of individual observation vectors $\mathcal{O}^{1}_{\tau=0}$ and $\mathcal{O}^{2}_{\tau=0}$ and a task result observation $\mathcal{O}^{\text{result}}_{\tau=0}$, is defined as: % <<< changed

\begin{equation}
\mathcal{O}^{\{1,2\}}_{\tau=0} =
\begin{bmatrix}
0 & 0 & 0 & 1 \\
0 & 0 & 0 & 1\\
0 & 0 & 1 & -
\end{bmatrix}, % <<< changed (aligned dims)
\label{observation}
\end{equation}

where each row corresponds to one robot’s observation vector 
$\mathcal{O}^{i}_{\tau=0} = [o^{i}_{\text{loc}}, o^{i}_{\text{hold}}, o^{i}_{\text{place}}, o^{i}_{\text{free}}]$. % <<< changed
A value of “1” in the last column indicates that both robots are currently in the $s^{i}_{\text{free}}$ state. 
This observation serves as the initial condition for belief inference within the IIBT-node, from which each agent begins reasoning about its subsequent actions based on $\mathcal{C}^{\{1,2\}}$ and $\Pi^{\{1,2\}}$. % <<< changed

At this initial stage, the value “1” in $\mathcal{O}^{\{1,2\}}_{\tau=0}$ indicates that both robots are in the $s^{i}_{\text{free}}$ state, signifying that the task execution has not yet begun. 
Under this condition, the IIBT-Node of $\mathcal{R}_1$ directs its reasoning focus toward the interactive node \textit{IsHolding(Bottle)}, while $\mathcal{R}_2$ remains inactive to prevent potential conflicts arising from parallel policy execution. 
To formalize this coordination logic, a joint extrinsic preference matrix $\mathcal{C}^{\{1,2\}}$ is defined as follows: % <<< changed

\begin{equation}
\mathcal{C}^1 =
\begin{bmatrix}
0 & 1 & 0 & 0 \\
0 & 0 & 0 & 1\\
1 & 0 & 0 & -
\end{bmatrix}.
\end{equation}

% \begin{equation}
% \mathcal{C}^{2} =
% \begin{bmatrix}
% 1 & 0 & 0 & 0 \\
% 1 & 2 & 0 & 0\\
% 1 & 0 & 0 & -
% \end{bmatrix}.
% \end{equation}

% \begin{equation}
% \mathcal{O}^{2} =
% \begin{bmatrix}
% 0 & 0 & 0 & 1 \\
% 0 & 0 & 0 & 1\\
% 0 & 0 & 1 & -
% \end{bmatrix}.
% \end{equation}

Each row in $\mathcal{C}^1$ corresponds to one robot’s preference distribution over the logical state set $\{l_{\text{loc}}, l_{\text{hold}}, l_{\text{place}}, l_{\text{free}}\}$. % <<< changed
The first row represents $\mathcal{R}_1$’s extrinsic preference, encouraging transition toward $s^{1}_{\text{hold}}$, while the second row expresses $\mathcal{R}_2$’s preference to remain “free.” 
During inference, $\mathcal{R}_1$ will thus select the \textit{pick(obj)} action only when $o^2_{\text{free}} = 1$, thereby establishing a logically consistent division of labor between the two robots. % <<< changed

However, the current observational state $\mathcal{O}^{\{1,2\}}_{\tau} \neq \{l^i_{\text{loc}}, l^i_{\text{free}}\}$ (Algorithm~\ref{alg:strategy_selection}, line 17), indicating that the precondition $l^i_{\text{loc}}$ for the \textit{pick(obj)} action is not yet satisfied. 
Consequently, $\mathcal{C}^i$ must be adaptively adjusted through logical evidence $\mathcal{L}^i$ to align the desired goal state with environmental constraints. 
This adaptive adjustment mechanism forms the foundation of interactive inference within the IIBT framework.

To satisfy the unsatisfied precondition $l^i_{\text{loc}}$ for \textit{pick(obj)}, 
the original $\mathcal{C}^i$ is augmented by a supplementary logical constraint matrix $\mathcal{L}^i$, introducing the necessary condition for spatial reachability: % <<< changed

\begin{equation}
\mathcal{L}^i =
\begin{bmatrix}
2 & 0 & 0 & 0\\
0 & 0 & 0 & 0\\
0 & 0 & 0 & -
\end{bmatrix}. % <<< changed
\end{equation}

The entry “2” in the first row explicitly encodes a higher-priority constraint on the \textit{moveTo(goal)} action. 
This adjustment ensures that, given the current observation, the agent prioritizes navigating to the goal location before attempting to grasp the object. 
Consequently, the updated comprehensive preference matrix becomes:

\[
\mathcal{C}^i = \mathcal{C}^i + \mathcal{L}^i =
\begin{bmatrix}
2 & 1 & 0 & 0 \\
0 & 0 & 0 & 1\\
1 & 0 & 0 & -
\end{bmatrix}. % <<< changed
\]

\begin{figure}[ht]
    \centering
    \includegraphics[width=1\textwidth]{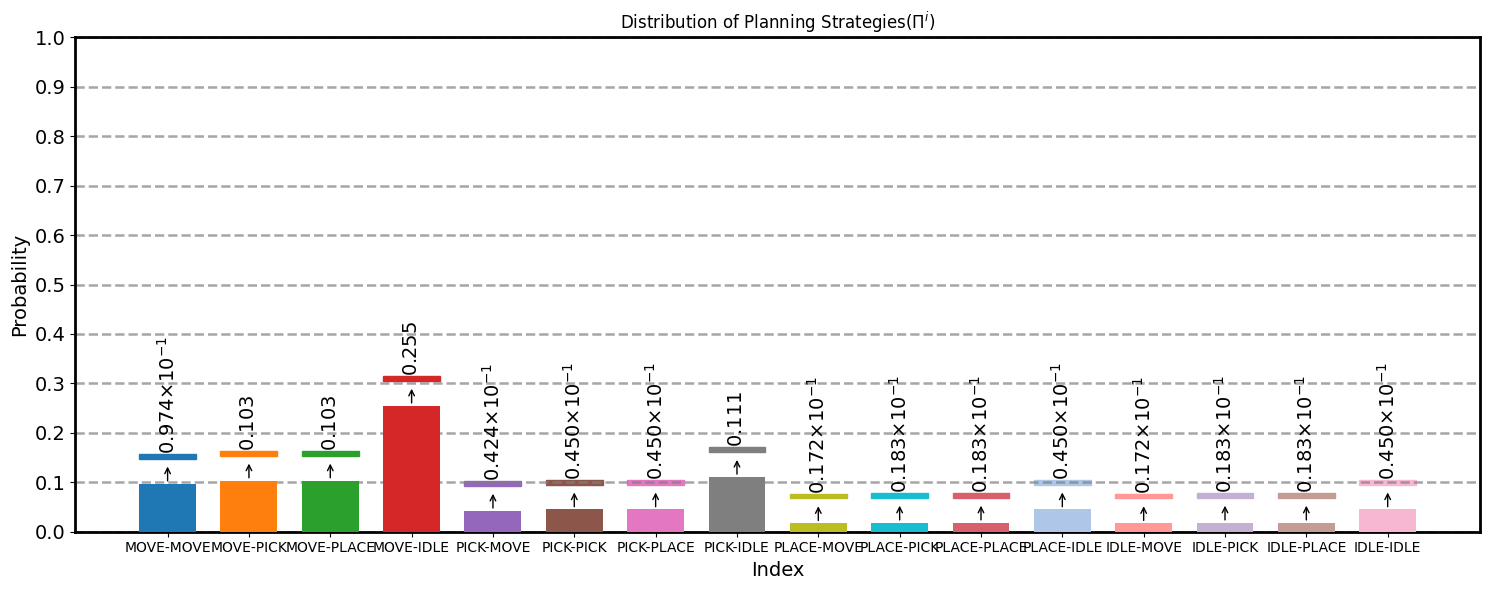}
    \caption{Probability distribution over joint strategies after EFE inference.}
    \label{fig:strategy_distribution}
\end{figure}

This update demonstrates that $\mathcal{C}^i$ evolves dynamically during task execution, incorporating contextual information and logical constraints derived from $\mathcal{O}^i_\tau$ and $\mathcal{L}^i$. 
Following Expected Free Energy (EFE) inference over $\Pi^{\{1,2\}} = \{\pi_0, \pi_1, ..., \pi_{15}\}$, the IIBT-node determines the MAP joint action as $\pi_{3} = (\text{moveTo(goal)}, \text{idle})$, with $\mathcal{P}(\pi_3)=0.255$ (Fig.~\ref{fig:strategy_distribution}). % <<< changed
This indicates that $\mathcal{R}_1$ navigates while $\mathcal{R}_2$ remains idle, achieving conflict-free coordination.

Upon completion of \textit{moveTo(goal)}, the temporary logical constraint $\mathcal{L}^i$ is removed:
\[
\mathcal{C}^i = \mathcal{C}^i - \mathcal{L}^i =
\begin{bmatrix}
0 & 1 & 0 & 0 \\
0 & 0 & 0 & 1\\
1 & 0 & 0 & -
\end{bmatrix}. % <<< changed
\]

\begin{figure}[ht]
\centering
\includegraphics[width=1\textwidth]{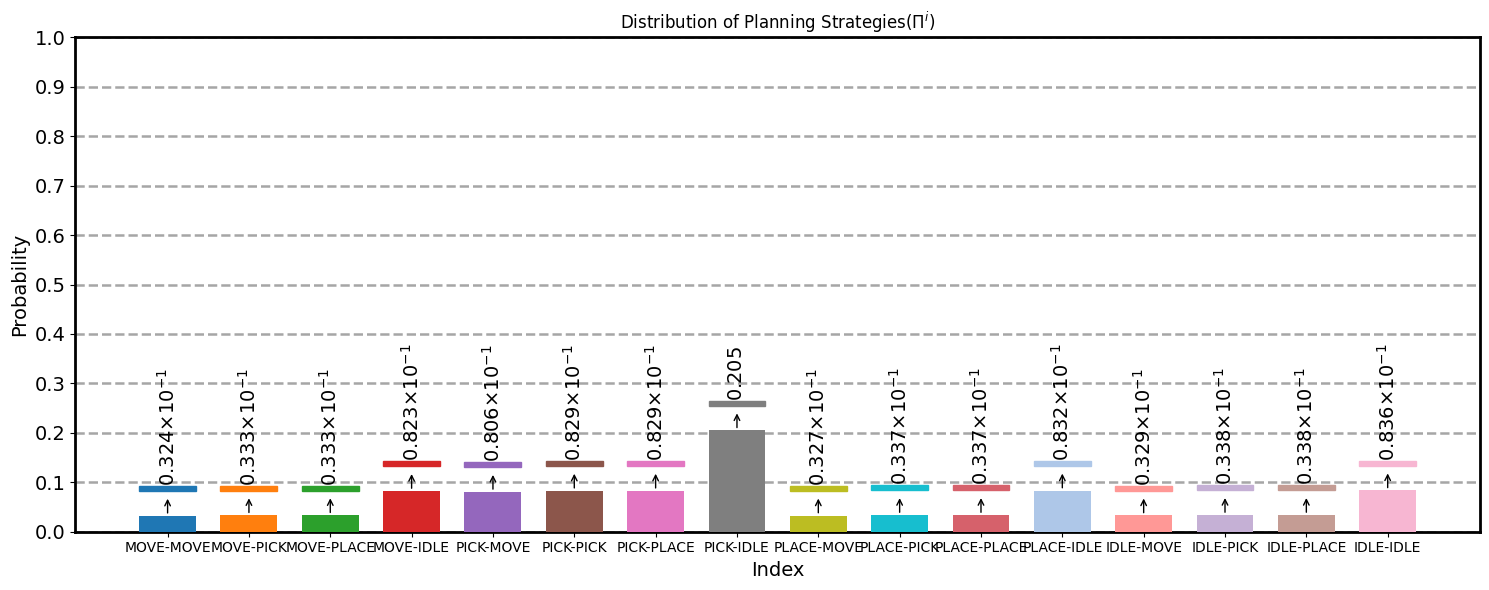}
\caption{Updated probability distribution over joint strategies after completion of the moveTo(goal) action.}
\label{fig:strategy_distribution_2}
\end{figure}

EFE inference is then repeated, yielding $\pi_7 = (\text{pick(obj)}, \text{idle})$ with $\mathcal{P}(\pi_7)=0.205$ (Fig.~\ref{fig:strategy_distribution_2}). 
This drives $\mathcal{R}_1$ to execute \textit{pick(obj)}, demonstrating dynamic re-evaluation and adaptive cooperation.

Next, robot $\mathcal{R}_2$ enters the interactive node \textit{IsHolding(Plate)}. 
It constructs its local constraint matrix $\mathcal{L}^2$, encoding both intrinsic preconditions and inter-agent requirements: % <<< changed

\begin{equation}
\mathcal{L}^2 =
\begin{bmatrix}
0 & 0 & 0 & 0 \\
0 & 2 & 0 & 0 \\
0 & 0 & 0 & -
\end{bmatrix}. % <<< changed
\end{equation}

\begin{figure}[ht]
    \centering
    \includegraphics[width=1\textwidth]{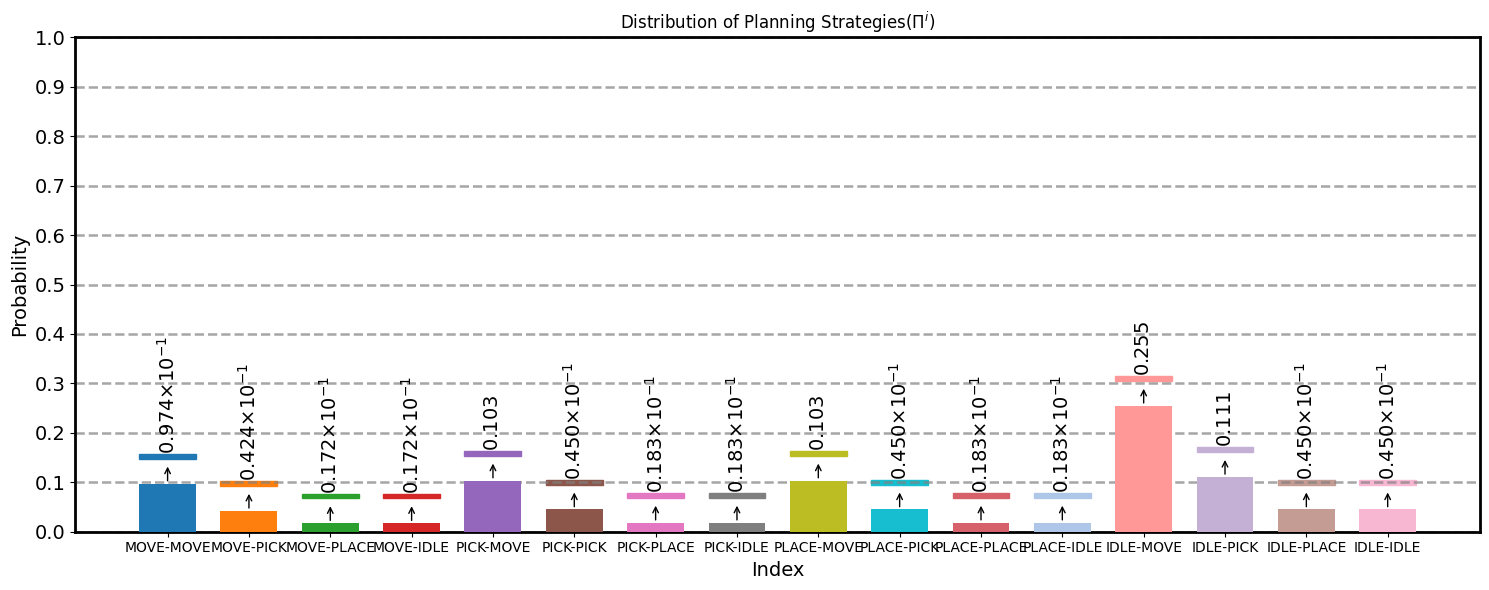}
    \caption{Probability distribution over joint strategies for $\mathcal{R}_2$ after completing \textit{moveTo(goal)}.}
    \label{fig:strategy_distribution_3}
\end{figure}

\begin{figure}[ht]
    \centering
    \includegraphics[width=1\textwidth]{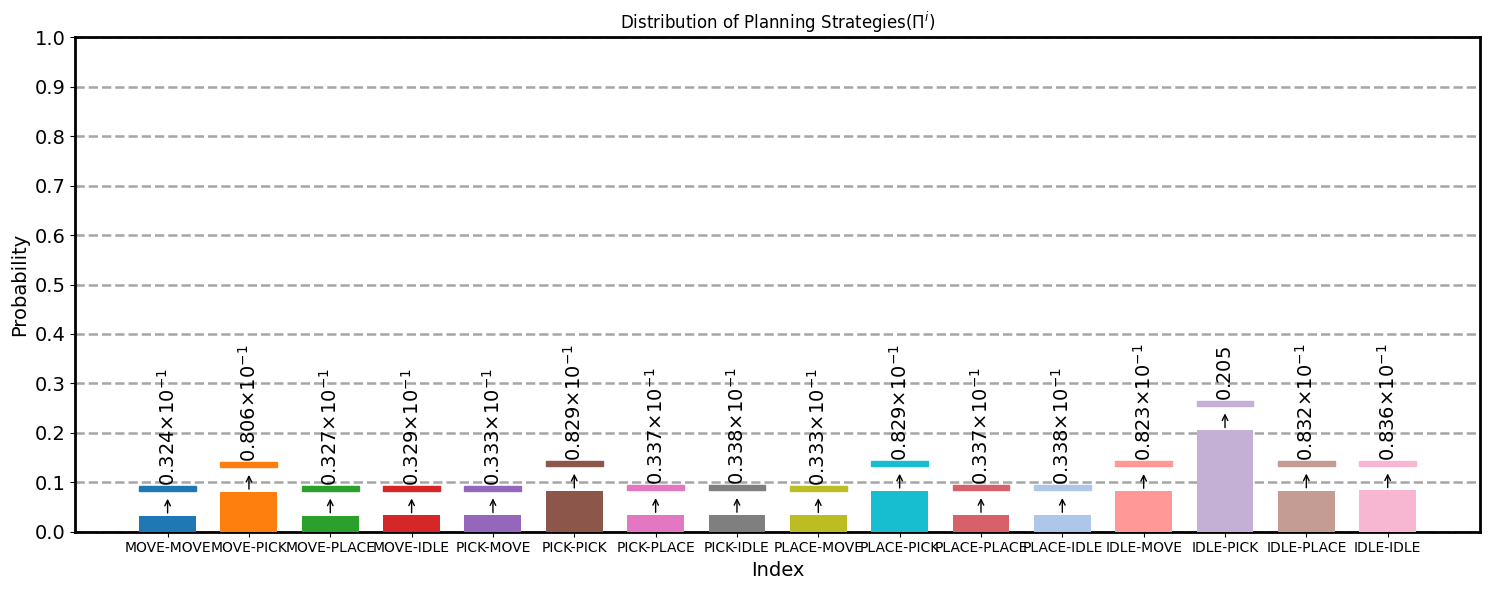}
    \caption{Updated probability distribution over joint strategies for $\mathcal{R}_2$ after \textit{pick(plate)}.}
    \label{fig:strategy_distribution_4}
\end{figure}

The first row imposes the dependency that $\mathcal{R}_1$ must be free before $\mathcal{R}_2$ executes \textit{pick(plate)}, while the second encodes $\mathcal{R}_2$’s intrinsic preference for holding. 
EFE inference over $\Pi^{\{1,2\}}$ produces distributions shown in Fig.~\ref{fig:strategy_distribution_3} and Fig.~\ref{fig:strategy_distribution_4}.

\begin{figure}[ht]
\centering
\includegraphics[width=1\textwidth]{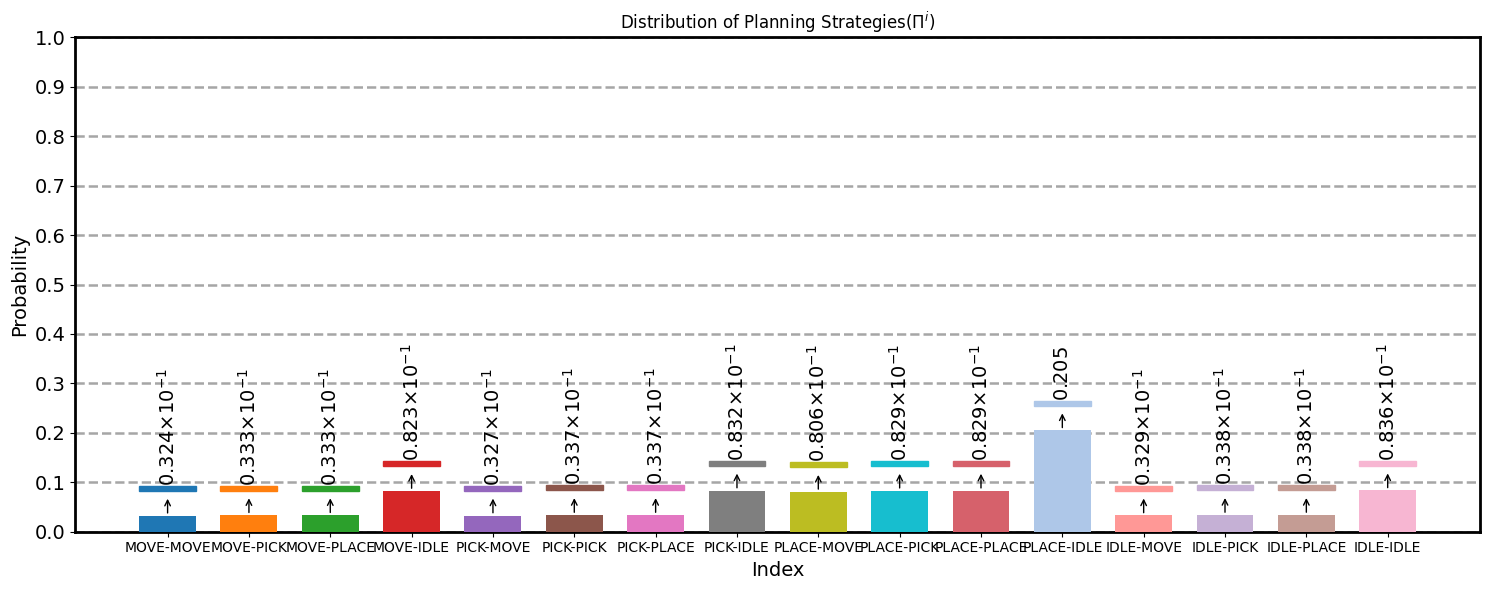}
\caption{Final probability distribution over joint strategies during the collaborative object placement phase.}
\label{fig:strategy_distribution_5}
\end{figure}

The selected joint strategies are 
$\pi_{12}=(\text{idle}, \text{moveTo(Goal B)})$ and $\pi_{13}=(\text{idle}, \text{pick(plate)})$, 
with $\mathcal{P}(\pi_{12})=0.255$ and $\mathcal{P}(\pi_{13})=0.205$. 
Afterward, both robots execute \textit{moveTo(Goal C)}. 
Upon arrival, $\mathcal{R}_1$ activates \textit{IsPlaced(Bottle,Plate)}, triggering another inference iteration to evaluate placement readiness (Fig.~\ref{fig:strategy_distribution_5}). % <<< changed

The final preference matrix $\mathcal{C}^{2}$ is updated as:
\[
\mathcal{C}^{2} =
\begin{bmatrix}
0 & 1 & 0 & 0\\
0 & 1 & 0 & 0\\
1 & 0 & 0 & -
\end{bmatrix}. % <<< changed
\]

Here, the first row denotes $\mathcal{R}_1$’s elevated preference for \textit{place(obj)}, while the second maintains $\mathcal{R}_2$’s preference for $s^{2}_{\text{free}}$, ensuring stability during placement. 
Through iterative inference and preference adjustment, the IIBT-Node effectively coordinates inter-robot dependencies and converges toward successful task completion. % <<< changed

\section{Conclusion}
\label{sec:conclusion}

This work presented the Interactive Inference Behavior Tree (IIBT-Node), a unified control node that integrates active inference with the modular architecture of Behavior Trees (BTs) for decentralized multi-robot cooperation. 
By embedding a dynamic preference matrix within each node, the proposed framework enables robots to infer, adapt, and coordinate their actions under uncertainty while preserving the interpretability and modularity of BTs. Extensive validation was conducted through both simulation and physical experiments using quadruped robots. In the simulated multi-robot navigation tasks, the IIBT-Node reduced the BT structural complexity by 76.2\%, while in real-world collaborative manipulation experiments, an equivalent reduction ranging from 70\% to 81.8\% was achieved. These results confirm that the proposed approach generalizes effectively across different action spaces, maintaining consistent reasoning performance and robust coordination under partial observability. Overall, the IIBT-Node provides a scalable and interpretable mechanism for multi-robot systems to achieve autonomous cooperation and conflict-free decision-making. Future work will focus on extending the framework to heterogeneous robot teams and exploring real-time learning of preference matrices in large-scale environments.

\bibliographystyle{IEEEtran}
% \balance
\bibliography{ref.bib}

@article{maisto2023interactive,
  title={Interactive inference: a multi-agent model of cooperative joint actions},
  author={Maisto, Domenico and Donnarumma, Francesco and Pezzulo, Giovanni},
  journal={IEEE Transactions on Systems, Man, and Cybernetics: Systems},
  year={2023},
  publisher={IEEE}
}

@article{pezzulo2024active,
  title={Active inference as a theory of sentient behavior},
  author={Pezzulo, Giovanni and Parr, Thomas and Friston, Karl},
  journal={Biological Psychology},
  pages={108741},
  year={2024},
  publisher={Elsevier}
}

@article{friston2023federated,
  title={Federated inference and belief sharing},
  author={Friston, Karl J and Parr, Thomas and Heins, Conor and Constant, Axel and Friedman, Daniel and Isomura, Takuya and Fields, Chris and Verbelen, Tim and Ramstead, Maxwell and Clippinger, John and others},
  journal={Neuroscience \& Biobehavioral Reviews},
  pages={105500},
  year={2023},
  publisher={Elsevier}
}

@article{lanillos2021active,
  title={Active inference in robotics and artificial agents: Survey and challenges},
  author={Lanillos, Pablo and Meo, Cristian and Pezzato, Corrado and Meera, Ajith Anil and Baioumy, Mohamed and Ohata, Wataru and Tschantz, Alexander and Millidge, Beren and Wisse, Martijn and Buckley, Christopher L and others},
  journal={arXiv preprint arXiv:2112.01871},
  year={2021}
}

@article{wirkuttis2021leading,
  title={Leading or following? Dyadic robot imitative interaction using the active inference framework},
  author={Wirkuttis, Nadine and Tani, Jun},
  journal={IEEE Robotics and Automation Letters},
  volume={6},
  number={3},
  pages={6024--6031},
  year={2021},
  publisher={IEEE}
}

@article{priorelli2024embodied,
  title={Embodied decisions as active inference},
  author={Priorelli, Matteo and Stoianov, Ivilin Peev and Pezzulo, Giovanni},
  journal={bioRxiv},
  pages={2024--05},
  year={2024},
  publisher={Cold Spring Harbor Laboratory}
}

@article{clodic2021implement,
  title={What Is It to Implement a Human-Robot Joint Action?},
  author={Clodic, Aurelie and Alami, Rachid},
  journal={Robotics, AI, and humanity: Science, ethics, and policy},
  pages={229--238},
  year={2021},
  publisher={Springer International Publishing}
}

@article{friedman2021active,
  title={Active inferants: An active inference framework for ant colony behavior},
  author={Friedman, Daniel Ari and Tschantz, Alec and Ramstead, Maxwell JD and Friston, Karl and Constant, Axel},
  journal={Frontiers in behavioral neuroscience},
  volume={15},
  pages={647732},
  year={2021},
  publisher={Frontiers Media SA}
}

@article{pezzato2023active,
  title={Active inference and behavior trees for reactive action planning and execution in robotics},
  author={Pezzato, Corrado and Corbato, Carlos Hern{\'a}ndez and Bonhof, Stefan and Wisse, Martijn},
  journal={IEEE Transactions on Robotics},
  volume={39},
  number={2},
  pages={1050--1069},
  year={2023},
  publisher={IEEE}
}

@inproceedings{colledanchise2019towards,
  title={Towards blended reactive planning and acting using behavior trees},
  author={Colledanchise, Michele and Almeida, Diogo and {\"O}gren, Petter},
  booktitle={2019 international conference on robotics and automation (ICRA)},
  pages={8839--8845},
  year={2019},
  organization={IEEE}
}

@article{li2024multi,
  title={Multi-Agent Dynamic Relational Reasoning for Social Robot Navigation},
  author={Li, Jiachen and Hua, Chuanbo and Ma, Hengbo and Park, Jinkyoo and Dax, Victoria and Kochenderfer, Mykel J},
  journal={arXiv preprint arXiv:2401.12275},
  year={2024}
}

@article{kim2000multi,
  title={Multi-agent systems: a survey from the robot-soccer perspective},
  author={Kim, Jong-Hwan and Vadakkepat, Prahlad},
  journal={Intelligent Automation \& Soft Computing},
  volume={6},
  number={1},
  pages={3--17},
  year={2000},
  publisher={Taylor \& Francis}
}

@misc{nixon1999process,
  title={Process control system using a control strategy implemented in a layered hierarchy of control modules},
  author={Nixon, Mark and Havekost, Robert B and Jundt, Larry O and Ott, Michael G and Webb, Arthur and Stevenson, Dennis and Lucas, Mike and Beoughter, Ken J},
  year={1999},
  month=jan # "~19",
  publisher={Google Patents},
  note={US Patent 5,862,052}
}

@article{luo2019importance,
  title={Importance sampling for online planning under uncertainty},
  author={Luo, Yuanfu and Bai, Haoyu and Hsu, David and Lee, Wee Sun},
  journal={The International Journal of Robotics Research},
  volume={38},
  number={2-3},
  pages={162--181},
  year={2019},
  publisher={SAGE Publications Sage UK: London, England}
}

@article{venkata2023kt,
  title={Kt-bt: A framework for knowledge transfer through behavior trees in multirobot systems},
  author={Venkata, Sanjay Sarma Oruganti and Parasuraman, Ramviyas and Pidaparti, Ramana},
  journal={IEEE Transactions on Robotics},
  year={2023},
  publisher={IEEE}
}

@article{gugliermo2023learning,
  title={Learning behavior trees from planning experts using decision tree and logic factorization},
  author={Gugliermo, Simona and Schaffernicht, Erik and Koniaris, Christos and Pecora, Federico},
  journal={IEEE Robotics and Automation Letters},
  volume={8},
  number={6},
  pages={3534--3541},
  year={2023},
  publisher={IEEE}
}

@inproceedings{hull2024communicating,
  title={Communicating Intent as Behaviour Trees for Decentralised Multi-Robot Coordination},
  author={Hull, Rhett and Moratuwage, Diluka and Scheide, Emily and Fitch, Robert and Best, Graeme},
  booktitle={2024 IEEE International Conference on Robotics and Automation (ICRA)},
  pages={7215--7221},
  year={2024},
  organization={IEEE}
}

@article{colledanchise2021implementation,
  title={On the implementation of behavior trees in robotics},
  author={Colledanchise, Michele and Natale, Lorenzo},
  journal={IEEE Robotics and Automation Letters},
  volume={6},
  number={3},
  pages={5929--5936},
  year={2021},
  publisher={IEEE}
}

@article{friston2024federated,
  title={Federated inference and belief sharing},
  author={Friston, Karl J and Parr, Thomas and Heins, Conor and Constant, Axel and Friedman, Daniel and Isomura, Takuya and Fields, Chris and Verbelen, Tim and Ramstead, Maxwell and Clippinger, John and others},
  journal={Neuroscience \& Biobehavioral Reviews},
  volume={156},
  pages={105500},
  year={2024},
  publisher={Elsevier}
}

@article{gugliermo2024evaluating,
  title={Evaluating behavior trees},
  author={Gugliermo, Simona and Dominguez, David Caceres and Iannotta, Marco and Stoyanov, Todor and Schaffernicht, Erik},
  journal={Robotics and Autonomous Systems},
  volume={178},
  pages={104714},
  year={2024},
  publisher={Elsevier}
}

@article{li2024embedding,
  title={Embedding multi-agent reinforcement learning into behavior trees with unexpected interruptions},
  author={Li, Xianglong and Li, Yuan and Zhang, Jieyuan and Xu, Xinhai and Liu, Donghong},
  journal={Complex \& Intelligent Systems},
  volume={10},
  number={3},
  pages={3273--3282},
  year={2024},
  publisher={Springer}
}

@article{liu2024autonomous,
  title={Autonomous Robot Task Execution in Flexible Manufacturing: Integrating PDDL and Behavior Trees in ARIAC 2023},
  author={Liu, Ruikai and Wan, Guangxi and Jiang, Maowei and Chen, Haojie and Zeng, Peng},
  journal={Biomimetics},
  volume={9},
  number={10},
  pages={612},
  year={2024},
  publisher={MDPI}
}

@article{bramblett2025implicit,
  title={Implicit Coordination using Active Epistemic Inference},
  author={Bramblett, Lauren and Reasoner, Jonathan and Bezzo, Nicola},
  journal={arXiv e-prints},
  pages={arXiv--2501},
  year={2025}
}

@article{scheide2025synthesizing,
  title={Synthesizing compact behavior trees for probabilistic robotics domains},
  author={Scheide, Emily and Best, Graeme and Hollinger, Geoffrey A},
  journal={Autonomous Robots},
  volume={49},
  number={1},
  pages={3},
  year={2025},
  publisher={Springer}
}

@article{wakayama2024active,
  title={Active inference in contextual multi-armed bandits for autonomous robotic exploration},
  author={Wakayama, Shohei and Candela, Alberto and Hayne, Paul and Ahmed, Nisar},
  journal={arXiv preprint arXiv:2408.04119},
  year={2024}
}

@article{blei2017variational,
  title={Variational inference: A review for statisticians},
  author={Blei, David M and Kucukelbir, Alp and McAuliffe, Jon D},
  journal={Journal of the American statistical Association},
  volume={112},
  number={518},
  pages={859--877},
  year={2017},
  publisher={Taylor \& Francis}
}

@article{jordan1999introduction,
  title={An introduction to variational methods for graphical models},
  author={Jordan, Michael I and Ghahramani, Zoubin and Jaakkola, Tommi S and Saul, Lawrence K},
  journal={Machine learning},
  volume={37},
  number={2},
  pages={183--233},
  year={1999},
  publisher={Springer}
}

@inproceedings{kingma2014auto,
  title={Auto-encoding variational Bayes},
  author={Kingma, Diederik P and Welling, Max},
  booktitle={International Conference on Learning Representations (ICLR)},
  year={2014}
}

@article{friston2010free,
  title={The free-energy principle: a unified brain theory?},
  author={Friston, Karl},
  journal={Nature reviews neuroscience},
  volume={11},
  number={2},
  pages={127--138},
  year={2010},
  publisher={Nature publishing group}
}

@article{parr2019generalised,
  title={Generalised free energy and active inference},
  author={Parr, Thomas and Friston, Karl J},
  journal={Biological cybernetics},
  volume={113},
  number={5},
  pages={495--513},
  year={2019},
  publisher={Springer}
}

@article{pezzulo2015active,
  title={Active inference, homeostatic regulation and adaptive behavioural control},
  author={Pezzulo, Giovanni and Rigoli, Francesco and Friston, Karl},
  journal={Progress in neurobiology},
  volume={134},
  pages={17--35},
  year={2015},
  publisher={Elsevier}
}

@article{friston2015active,
  title={Active inference and learning},
  author={Friston, Karl and FitzGerald, Thomas and Rigoli, Francesco and Schwartenbeck, Philipp and Pezzulo, Giovanni and others},
  journal={Neuroscience \& Biobehavioral Reviews},
  volume={68},
  pages={862--879},
  year={2016},
  publisher={Elsevier}
}

@article{schwartenbeck2019computational,
  title={Computational mechanisms of curiosity and goal-directed exploration},
  author={Schwartenbeck, Philipp and Passecker, Johannes and Hauser, Tobias U and FitzGerald, Thomas HB and Kronbichler, Martin and Friston, Karl J},
  journal={elife},
  volume={8},
  pages={e41703},
  year={2019},
  publisher={eLife Sciences Publications, Ltd}
}

\end{document}